\documentclass[manuscript,screen]{acmart}

\usepackage{algorithmic}
\usepackage{array}
\usepackage[caption=false,font=normalsize,labelfont=sf,textfont=sf]{subfig}
\usepackage{textcomp}
\usepackage{stfloats}
\usepackage{url}
\usepackage{verbatim}
\usepackage{graphicx}

\usepackage{hyperref}
\hypersetup{
  hidelinks
}

\usepackage{multirow,bbding}
\usepackage{amsmath,amsfonts,bm}
\usepackage{threeparttable}

\usepackage{booktabs}
\usepackage{wrapfig}

\usepackage{tcolorbox}
\usepackage{mdframed}
\usepackage{wasysym}
\usepackage{tablefootnote}
\usepackage{longtable}

\usepackage{fontawesome5}
\usepackage{tikz}
\usepackage{balance}

\AtBeginDocument{%
  }

\setcopyright{acmlicensed}
\acmDOI{XXXXXXX.XXXXXXX}

\acmISBN{978-1-4503-XXXX-X/18/06}

\begin{document}
\tikzstyle{box} = [rectangle, minimum width=3cm, minimum height=1cm, text centered, draw=black]
\tikzstyle{arrow} = [thick,->,>=stealth]

\usetikzlibrary{positioning}

\title{Networking Systems for Video Anomaly Detection: A Tutorial and Survey}

\author{Jing Liu}
\email{jingliu19@fudan.edu.cn}
\orcid{0000-0002-2819-0200}

\affiliation{%
  \institution{Fudan University}
  \department{School of Information Science and Technology}
  \streetaddress{220 Handan Road}
  \city{Shanghai}
  \postcode{200433}
  \country{China}
}
\affiliation{%
  \institution{The University of British Columbia}
  \department{Department of Electrical and Computer Engineering}
  \streetaddress{2329 West Mall}
  \city{Vancouver}
  \state{British Columbia}
  \postcode{V6T 1Z4}
  \country{Canada}
}
\affiliation{%
  \institution{Duke Kunshan University}
  \department{Division of Natural and Applied Sciences}
  \streetaddress{8 Duke Avenue}
  \city{Kunshan}
  \state{Jiangsu Province}
  \postcode{215316}
  \country{China}
}

\author{Yang Liu}
\authornote{Corresponding authors.}
\email{yangliu@cs.toronto.edu}
\orcid{0000-0002-1312-0146}
\affiliation{%
  \institution{Soochow University}
  \department{School of Future Science and Engineering}
  \streetaddress{1 Jiuyong West Road, Wujiang District}
  \city{Suzhou}
  \state{Jiangsu Province}
  \postcode{215222}
  \country{China}
}

\author{Jieyu Lin}
\email{jieyu.lin@mail.utoronto.ca}
\orcid{0000-0002-6875-658X}
\affiliation{%
  \institution{University of Toronto}
  \department{Department of Electrical and Computer Engineering}
  \streetaddress{27 King's College Circle}
  \city{Toronto}
  \state{Ontario}
  \postcode{M5S 1A1}
  \country{Canada}
}

\author{Jielin Li}
\email{jielinli@connect.hku.hk}
\orcid{0009-0004-7728-2461}
\affiliation{%
  \institution{The University of Hong Kong}
  \department{Department of Computer Science}
  \streetaddress{Pokfulam Road}
  \city{Hong Kong}
  \country{Hong Kong SAR}
}

\author{Liang Cao}
\email{liangcao@mit.edu}
\orcid{0000-0002-2880-3097}
\affiliation{%
  \institution{Massachusetts Institute of Technology}
  \department{Department of Chemical Engineering}
  \streetaddress{77 Massachusetts Avenue}
  \city{Cambridge}
  \state{Massachusetts}
  \postcode{02139}
  \country{United States}
}

\author{Peng Sun}
\authornotemark[1]
\email{peng.sun568@duke.edu}
\orcid{0000-0003-1829-5671}
\affiliation{%
  \institution{Duke Kunshan University}
  \department{Division of Natural and Applied Sciences}
  \streetaddress{8 Duke Avenue}
  \city{Kunshan}
  \state{Jiangsu Province}
  \postcode{215316}
  \country{China}
}

\author{Bo Hu}
\authornotemark[1]
\email{bohu@fudan.edu.cn}
\orcid{0000-0001-6348-010X}
\affiliation{%
  \institution{Fudan University}
  \department{School of Information Science and Technology}
  \streetaddress{220 Handan Road}
  \city{Shanghai}
  \postcode{200433}
  \country{China}
}

\author{Liang Song}
\authornotemark[1]
\email{songl@fudan.edu.cn}
\orcid{0000-0002-8143-9052}
\affiliation{%
  \institution{Fudan University}
  \department{Academy for Engineering \& Technology}
  \streetaddress{220 Handan Road}
  \city{Shanghai}
  \postcode{200433}
  \country{China}
}

\author{Azzedine Boukerche}
\email{aboukerc@uOttawa.ca}
\orcid{0000-0002-3851-9938}
\affiliation{%
  \institution{University of Ottawa}
  \department{School of Electrical Engineering and Computer Science}
  \streetaddress{75 Laurier Avenue East}
  \city{Ottawa}
  \state{Ontario}
  \postcode{K1N 6N5}
  \country{Canada}
}

\author{Victor C.M. Leung}
\email{vleung@ece.ubc.ca}
\orcid{0000-0003-3529-2640}
\affiliation{%
  \institution{Shenzhen MSU-BIT University}
  \department{Artificial Intelligence Research institute}
  \streetaddress{1 International University Park Road, Dayun New Town}
  \city{Shenzhen}
  \state{Guangdong Province}
  \postcode{518172}
  \country{China}
}
\affiliation{%
  \institution{Shenzhen University}
  \department{College of Computer Science and Software Engineering}
  \streetaddress{3688 Nanhai Avenue}
  \city{Shenzhen}
  \state{Guangdong Province}
  \postcode{518060}
  \country{China}
}
\affiliation{%
  \institution{The University of British Columbia}
  \department{Department of Electrical and Computer Engineering}
  \streetaddress{2329 West Mall}
  \city{Vancouver}
  \state{British Columbia}
  \postcode{V6T 1Z4}
  \country{Canada}
}

\renewcommand{\shortauthors}{Jing Liu et al.}

\begin{abstract}
  The increasing utilization of surveillance cameras in smart cities, coupled with the surge of online video applications, has heightened concerns regarding public security and privacy protection, which propelled automated Video Anomaly Detection (VAD) into a fundamental research task within the Artificial Intelligence (AI) community. With the advancements in deep learning and edge computing, VAD has made significant progress and advances synergized with emerging applications in smart cities and video internet, which has moved beyond the conventional research scope of algorithm engineering to deployable Networking Systems for VAD (NSVAD), a practical hotspot for intersection exploration in the AI, IoVT, and computing fields. In this article, we delineate the foundational assumptions, learning frameworks, and applicable scenarios of various deep learning-driven VAD routes, offering an exhaustive tutorial for novices in NSVAD. In addition, this article elucidates core concepts by reviewing recent advances and typical solutions and aggregating available research resources accessible at \url{https://github.com/fdjingliu/NSVAD}. Lastly, this article projects future development trends and discusses how the integration of AI and computing technologies can address existing research challenges and promote open opportunities, serving as an insightful guide for prospective researchers and engineers.
\end{abstract}



\maketitle

\section{Introduction}~\label{sec1}
As one of the core technologies of the ubiquitous Internet of Video Things (IoVT), Video Anomaly Detection (VAD) aims to use video sensors to automatically discover unexpected spatial-temporal patterns and detect unusual events that may cause security problems or economic losses, such as traffic accidents, violent behaviors, and offending contents \cite{liu2024generalized}. With the widespread use of surveillance cameras in smart cities \cite{ramachandra2020survey} and the boom of online video applications powered by 4/5G communication technologies, traditional human inspection is no longer able to accurately monitor the video data generated around the clock, which is not only time-consuming and labor-intensive but also poses the risk of leaking important information (e.g., biometrics and sensitive speech). In contrast, VAD-empowered IoVT applications \cite{jedari2020video}, such as Intelligent Video Surveillance Systems (IVSS) and automated content analysis platforms, can process massive video streams online and detect events of interest in real-time, sending only noteworthy anomaly parts for human review, significantly reducing data storage and communication costs, and helping to eliminate public concerns about data security and privacy protection. As a result, VAD has gained widespread attention in academia and industry over the last decade and has been used in emerging fields \cite{cao2024real,peng2025defective,cao2025adaptive} such as information forensics \cite{yu2023video}, industrial manufacturing \cite{LIU2024123718,wang2024mgr3net} in smart cities as well as online content analysis in mobile video applications \cite{yin2024systematic}.

VAD extends the data scope of conventional Anomaly Detection (AD) from time series, images, and graphs to video, which not only needs to cope with the endogenous data complexity, but also needs to take into account the computational and communication costs in resource-limited devices \cite{jiang2021survey}. Specifically, the inherent high-dimensional structure of video data, high information density and redundancy, heterogeneity of temporal and spatial patterns, and feature entanglement between foreground targets and background scenes make VAD more challenging than traditional AD tasks at the levels of representation learning and anomaly discrimination \cite{mishra2024skeletal}. Existing studies \cite{FFP, amp, HSNBM,li2024cross} have shown that high-performance VAD models need to target the modeling of appearance and motion information, i.e., the difference between regular events and anomalous examples in both spatial and temporal dimensions. In contrast to time series AD which mainly measures periodic temporal patterns of variables, and image AD which only focuses on spatial contextual deviations, VAD needs to extract both discriminative spatial and temporal features from a large amount of redundant information (e.g., repetitive temporal contexts and label-independent data shifts), as well as to learn the differences between normal and anomalous events in terms of the local appearances and global motions \cite{qiu2024video,liu2025domain,liu2022multilevel}. 

However, video anomalies are ambiguous and subjective \cite{liu2021onedimensional,liu2023dsdcla}. The same driving behavior can be classified differently depending on road conditions and contextual environments. For example, riding a horse in a grassland is usually normal, whereas a horse appearing on a highway would be considered an anomaly. On the one hand, compared to regular events, anomalies in the real world are difficult to be comprehensively predefined and have a much lower overall frequency of occurrence, making them difficult to collect. Labeling a sufficient number of all possible abnormal samples for model training is almost impossible. As a result, traditional supervised learning-based classification models are usually ineffective in dealing with AD tasks \cite{mavskova2024deep}. On the other hand, since video storage and transmission costs are significantly higher than other data modalities, engineers favor processing such data on the end or edge side to reduce communication overhead. As we all know, such devices, including surveillance cameras, smartphones, and local servers, are computationally resource-limited. Therefore, it is highly practical to develop deployable VAD systems for real-world applications, which requires concerted efforts by AI beyond communities.

\begin{figure}[t]
  \centering
  \includegraphics[width=.98\linewidth]{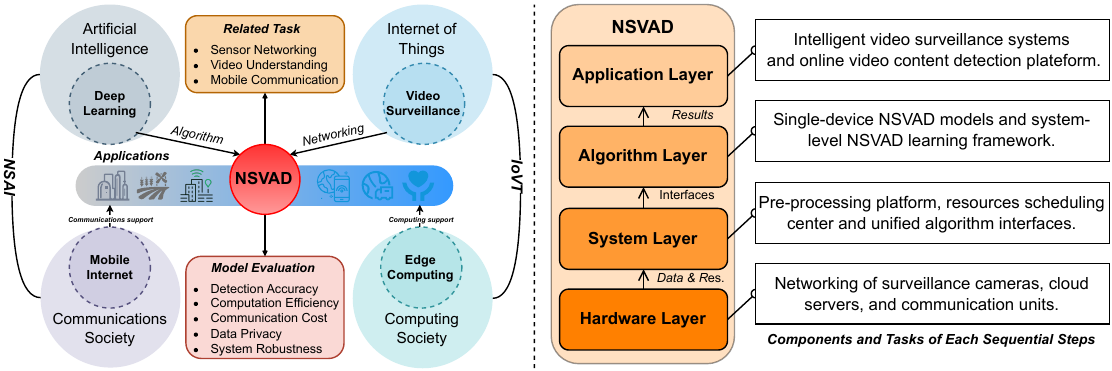}
  \caption{Topology diagram of research scope of NSAVD (Left) and its key sequential steps (Right).}
  \label{nsvad}
\end{figure}

In this article, we extend the conventional scope of VAD from algorithm engineering on spatial-temporal anomaly detection to practical research towards real-world applications, termed \textbf{N}etworking \textbf{S}ystems for \textbf{V}ideo \textbf{A}nomaly \textbf{D}etection (NSVAD), to engage a broader readership from the IoT and computing communities. According to research objectives and involved domains, NSVAD is delineated into the hierarchical architecture shown in Fig.~\ref{nsvad}, encompassing: 1) Hardware Layer, consisting of various video sensors, communication units, and computing servers, etc, responsible for data acquisition, transmission, processing, and result reporting, as well as device networking; 2) System Layer, targeting resource optimization and algorithm deployment platforms for large-scale IoT applications, linking the Hardware Layer and the Algorithm Layer, supporting the configurable deployment of VAD tasks on various terminals; 3) Algorithm Layer, focusing on the development of detection algorithms and scene-specific models driven by artificial intelligence, especially deep learning; 4) Application Layer, encompassing IVSS in modern factories, agriculture, and smart cities, as well as various online video applications powered mobile internet. Most existing works belong to the Algorithm layer and solely concentrate on VAD model design, overlooking resource costs and challenges in real-world scenarios. For large-scale IoT and mobile video internet, NSVAD with stable performance and reasonable overheads that support online detection necessitates sensor networking research and support for resource optimization from computing communities. We review the recent advancements and typical methods in the algorithm layer and provide our latest explorations in the system layer to inspire readers to develop NSVAD toward real-world scenarios.

Thanks to the development of edge AI \cite{walia2023ai} and artificial neural networks \cite{li2021survey}, deep learning-driven NSVAD algorithms have made significant progress in recent years and derived Unsupervised (UVAD), Weakly-supervised (WsVAD), and Fully-unsupervised (FuVAD) routes \cite{liu2024generalized}. They have liberated human beings from massive videos analysis works and alleviated public information security concerns. Compared to early manual feature engineering, deep architectures such as convolutional neural networks and attention mechanisms can extract spatial-temporal representations from video sequences end-to-end without applying human a priori, empowering IVSS to process videos in different resolutions and scenarios. Therefore, researchers in this field are currently focusing on deep structure design and optimization learning strategies. They have creatively proposed multimodal VAD \cite{HL-Net, HL-Net+, RTFM}, Open-Set AVD (OSVAD) \cite{UBnormal,osvad}, Open-Vocabulary VAD (OVVAD) \cite{ovvad}, video anomaly segmentation \cite{tian2024latency}, and anomaly retrieval \cite{wu2024toward} tasks as well as integrated detection systems that can be deployed in practical scenes, such as modern manufacturing \cite{amp}, smart city \cite{crc}, and automated driving \cite{ROADMAP}. In addition to algorithm design, researchers from Networking Systems of AI (i.e., research on the deep convergence of communication and AI) \cite{nsai} and IoVT (i.e., subfields of IoT focusing on video sensor design, networking, and data processing) \cite{liu2021surveya} have begun to explore the design deployment-oriented VAD systems to collaboratively deal with multiple challenges (e.g., multi-view cross-scene heterogeneous videos and communication overheads) that come from the dynamic scenarios at the application layer and the limited resources at the hardware layer. These explorations and progress greatly expand the research boundaries and application scenarios of VAD, promoting it as an intelligent system science, i.e., NSVAD.

Although there have been some reviews \cite{pang2021deep,cook2019anomaly,santhosh2020anomaly,samaila2024video} focusing on AD and combing its related work, due to the limited research horizon, the early works usually regard VAD as a fringe research task in the AD community. They focus on time series or images but lack an illuminating analysis of the AD task in video data. Recent survey papers \cite{nayak2021comprehensive,ramachandra2020survey,chandrakala2022anomaly} continue to focus on unsupervised NSVAD routes in the same vein as the conventional AD task, i.e., using only normal samples to train generative models to learn the prototypical patterns of regular events and to discriminate uncharacterizable test samples as anomalies \cite{FFPN}. Such reviews, while providing a comprehensive taxonomy of VAD research from the outlier detection perspective, have been informative in the last decade when unsupervised methods have dominated the NSVAD algorithm research. However, they ignore the emerging weakly-supervised \cite{MIST, RTFM} and fully unsupervised \cite{SDOR, GCL} routes, which are of limited value in guiding further research.

Considering the differences in knowledge bases and orientations of readers, this article provides an in-depth analysis of the basic concepts, related knowledge, and recent advances involved in NSVAD research and summarizes the available research resources. We systematically analyze the assumptions, frameworks, scenarios, advantages, and disadvantages of unsupervised, weakly supervised, and fully unsupervised VAD routes and explain in detail the relevant domain knowledge involved in each route. In addition, we introduce our NSVAD systems designed for dynamic environments in complex scenarios such as industrial IoT \cite{cao2025novel} and smart cities \cite{liu2025crcl} to guide the NSVAD research in specific applications. 
Finally, we forecast the research challenges, trends, and possible opportunities to inspire future exploration.

\subsection{Attention Analysis}

\begin{figure}[t]
  \centering
  \includegraphics[width=.95\linewidth]{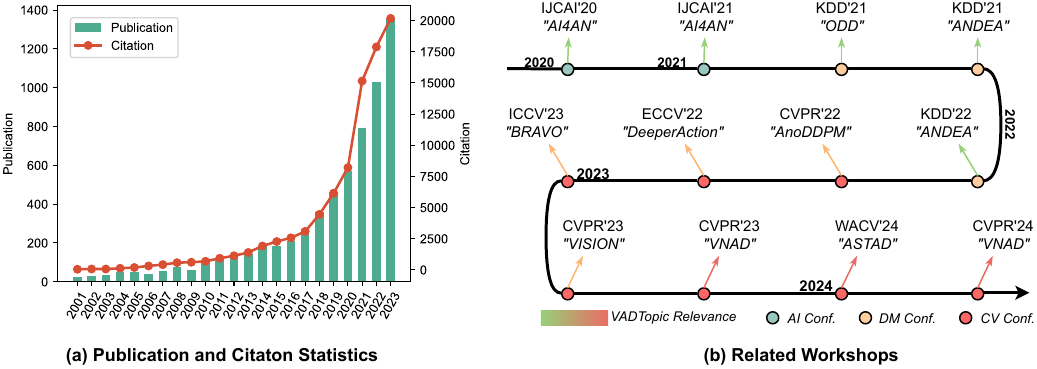}
  \caption{Research hotness analysis. We count \textbf{(a)} the number of NSVAD-related publications and their citations in the past 23 years and organize \textbf{(b)} the AD-related workshops in conferences on Artificial Intelligence (AI), Data Mining (DM), and Computer Vision (CV).}
  \label{rdfx}
\end{figure}

We searched the number of publications and citations with the topic of \textit{"video anomaly detection"} in various mainstream academic databases (e.g., IEEE Xplore, ACM Digital Library, SpringerLink, ScienceDirect, and DBLP) to quantitatively present the research hotness, as shown in Fig.~\ref{rdfx}(a). Early VAD works were limited by hand-crafted features, which cannot handle complicated videos and rely on human a priori, thus having a slow start. Encouraged by the remarkable success of deep learning in video understanding tasks (e.g., action recognition, scene understanding, expression recognition \cite{wang2022dpcnet} and multimodal perception \cite{yang2022emotion}), VAD research saw a boom after 2010, with an explosive growth in the number of publications and citations that continues to the present. On the one hand, the spread of surveillance cameras and streaming media platforms has provided sufficient data support for NSVAD research, making it possible to train large-scale deep models with high-performance GPUs. On the other hand, the increasing demand for offending video content detection in various scenarios drives many researchers and engineers from AI and IoT fields to devote themselves to NSVAD research. 

Fig.~\ref{rdfx}(b) presents the AD-related workshops in the top computer science conferences in the past four years. The changes in data types and application scenarios in these workshops show that AD tasks on visual data, especially videos, have dominated the community. For example, the \textit{"DeeperAction"} workshop explicitly identified anomalous behavior recognition in surveillance videos as the following research hotspot in behavior analysis. The first \textit{"ASTAD"} workshop at WACV'24 centered on anomalous detection of spatial-temporal data and its application to computer vision tasks. In addition, the latest workshops, \textit{"BRAVO"} and \textit{"VISION"}, explored the application of AD technology in areas such as autonomous driving and modern manufacturing, further demonstrating the high hotness and broad application prospects of VAD. To make it easier for beginners to navigate these workshops, we have categorized all the available resources, please see our public GitHub repository at \url{https://github.com/fdjingliu/NSVAD}.

\subsection{Related Work}
\begin{table}[t]
  \centering
  \caption{Comparison with related survey papers and conference tutorials.}
  \label{tab:related}
\begin{threeparttable}
  \resizebox{.98\textwidth}{!}{
  \begin{tabular}{@{}cclcccccccccc@{}}
  \toprule
  \multirow{2}{*}{\textbf{Year}} & \multirow{2}{*}{\textbf{Ref.}} & \multicolumn{1}{c}{\multirow{2}{*}{\textbf{Perspective \& Main Focus}}} & \multicolumn{2}{c}{\textbf{Type}} & \multicolumn{5}{c}{\textbf{Research Routes \& Open Task}} & \multicolumn{3}{c}{\textbf{Content Analysis}}\\ \cmidrule(l){4-5} \cmidrule(l){6-10} \cmidrule(l){11-13}  
  & & \multicolumn{1}{c}{}& Survey & Tutorial& UVAD  & WsVAD & FuVAD & OSVAD & OVVAD & Advances Review & Basic Knowledge & Practical Cases \\ \midrule
 2018& \cite{kiran2018overview} & Un- and semi-supervised VAD&\Checkmark &  &  \CIRCLE&  \Circle&  \LEFTcircle&  \Circle&  \Circle&  \CIRCLE &  \Circle &  \Circle \\
 2019& \cite{cook2019anomaly} & Time series AD in IoT  &\Checkmark &  & \astrosun & \astrosun & \astrosun &  \astrosun&  \astrosun&  \CIRCLE &  \CIRCLE &  \Circle \\
 2020& \cite{santhosh2020anomaly} & VAD in traffic scene&\Checkmark &  &  \CIRCLE&  \LEFTcircle&  \Circle&  \Circle&  \Circle&  \CIRCLE &  \Circle &  \Circle \\
 2022& \cite{ramachandra2020survey} & Unsupervised VAD in single-scene&\Checkmark &  &  \CIRCLE&  \LEFTcircle&  \LEFTcircle&  \Circle&  \Circle&  \CIRCLE &  \CIRCLE &  \Circle \\
 2021& \cite{nayak2021comprehensive} & Deep learning-based unsupervised VAD&\Checkmark &  &  \CIRCLE&  \Circle&  \Circle&  \Circle&  \Circle&  \CIRCLE &  \Circle &  \Circle \\
 2021& \cite{rezaee2021survey} & Unsupervised VAD in crowd sence&\Checkmark &  &  \CIRCLE&  \Circle&  \Circle&  \Circle&  \Circle&  \CIRCLE &  \Circle &  \Circle \\
 2021& \cite{pang2021deep} & Deep learing-based unsupervised AD&\Checkmark &  & \astrosun  & \astrosun & \astrosun &  \astrosun&  \astrosun&  \CIRCLE &  \CIRCLE &  \Circle \\
 2021& \cite{blazquez2021review} & Time series AD&\Checkmark &  & \astrosun & \astrosun & \astrosun &  \astrosun&  \astrosun&  \CIRCLE &  \Circle &  \Circle \\
 2022& \cite{chandrakala2022anomaly} & Unsupervised and supervised VAD&\Checkmark &  &  \CIRCLE&  \LEFTcircle&  \Circle&  \Circle&  \Circle&  \CIRCLE &  \Circle &  \Circle \\
 2022& \cite{raja2022analysis} & Unsupervised VAD  &\Checkmark &  &  \CIRCLE&  \CIRCLE&  \Circle&  \Circle&  \Circle&  \CIRCLE &  \Circle &  \Circle \\
 2023& - & Un- and weakly-supervised VAD&  &\Checkmark &  \CIRCLE&  \CIRCLE&  \Circle&  \Circle&  \Circle&  \CIRCLE &  \LEFTcircle &  \Circle \\
 2024& \cite{liu2024generalized} & Generalized VAD&\Checkmark &  &  \CIRCLE&  \CIRCLE&  \CIRCLE&  \LEFTcircle&  \Circle&  \CIRCLE &  \Circle &  \Circle \\
 2024& Ours & NSVAD routes in video IoT&  &\Checkmark &  \CIRCLE&  \CIRCLE&  \CIRCLE&  \CIRCLE&  \CIRCLE&  \CIRCLE &  \CIRCLE &  \CIRCLE \\ \bottomrule
  \end{tabular}}
  \begin{tablenotes}
\footnotesize
\item \CIRCLE: Systematic compendium and presentation to the routes/tasks/contents. \LEFTcircle: Briefly mentioned. \Circle: Not presented. \astrosun: Not applicable.
  \end{tablenotes} 
\end{threeparttable}
  \end{table}

To our knowledge, this is the first tutorial-type paper on NSVAD, providing a systematic overview of the basics, recent advances, and practical applications of various VAD routes. Previous papers \cite{liu2024generalized,ramachandra2020survey,chandrakala2022anomaly,kiran2018overview} primarily focus on literature reviews, while conference tutorials lack systematic content, as shown in Table~\ref{tab:related}. Considering non-specialists' limited background knowledge, this article emphasizes clear explanations of basic concepts and models. We introduce task definitions and learning frameworks for unsupervised, weakly supervised, and fully unsupervised VAD, as well as emerging tasks like open-set \cite{osvad}, open vocabulary  \cite{ovvad}, and glance VAD \cite{zhang2024glancevad}.

Initially, VAD was considered a fringe topic in the broader AD community and only briefly mentioned in AD overviews \cite{cook2019anomaly,pang2021deep}, lacking comprehensive surveys. Recent efforts have started organizing the state of VAD research \cite{ramachandra2020survey,chandrakala2022anomaly,raja2022analysis,nayak2024comprehensive}, but they focus mainly on unsupervised methods and overlook rising research routes like WsVAD \cite{MIR} and FuVAD \cite{HL-Net}. These routes offer reliable performance in real-world applications, and their importance is increasingly recognized. Fully unsupervised methods, in particular, allow efficient VAD model learning from large-scale video streams. Recent works, such as \cite{liu2024generalized}, summarize these developments but do not cover emerging tasks like OSVAD \cite{UBnormal,osvad} and OVVAD \cite{ovvad}, which are of crucial value for IoT applications.

Researchers in the AD community have paid attention to the continued progress and application prospects of VAD and introduced it to attendees through conference tutorials. For example, Pang \textit{\textit{et al.}}\footnote{\url{https://cvpr.thecvf.com/virtual/2023/tutorial/18560}} organized a tutorial titled \textit{"Recent Advances in Anomaly Detection"} at CVPR2023, focusing on recent work on deep learning-driven unsupervised and weakly supervised VAD. We think this is a good start and highly appreciate the contribution of the organizers to the field. However, a more unified conceptual statement, more comprehensive documentation, and more systematic analysis of challenges and opportunities are necessary to inspire a wider community of readers.
\begin{figure}[t]
  \centering
  \includegraphics[width=.8\linewidth]{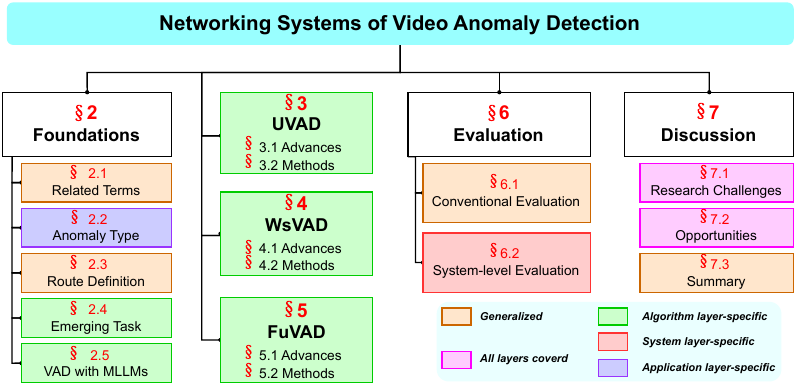}
  \caption{Content navigation of this article. 
 }
  \label{section}
\end{figure}

\subsection{Contribution Summary}
Given that NSVAD research has become an explicit hotspot in AI, computing, and IoT communities, and shows great potential for applications in emerging scenarios such as smart cities and mobile internet, we aim to provide systematic and inspiring guidance. This article is aimed at researchers and engineers who understand the main concepts and basic knowledge of AI but have no experience in NSVAD. We provide a comprehensive statement of unsupervised, weakly supervised, and fully unsupervised VAD routes, as well as various types of emerging tasks to satisfy readers with different backgrounds and needs. The contributions of this article can be summarized in the following four points:
\begin{itemize}
  \item To the best of our knowledge, this article will be the first tutorial-type paper focusing on Networking Systems for Video Anomaly detection, which not only provides a well-structured guide for non-specialized readers but also promises to bring together researchers from AI, IoT, and computing societies to promote NSVAD research.
  \item Focusing on AD in IoVT from the NSAI perspective, we comprehensively sort out the UVAD, WsVAD, and FuVAD routes and state their basic assumptions, learning paradigms, and applicability scenarios of each scheme.
  \item We open source available resources (e.g., benchmark datasets, code bases, literature, workshops, and tutorials) and provide our studies on NSAVD in industry and smart cities. 
  \item We analyze the development sequence between various research routes by empirically reviewing the recent advances and discussing the future vision of NSVAD in the context of trends and concerns in NSAI and IoVT.
\end{itemize}

\subsection{Section Navigation}

Based on the relationship between the content of the individual sections and the NSVAD architecture shown in Fig.~\ref{nsvad}, we organize the remainder of this article and provide an intuitive navigation map shown in Fig.~\ref{section}. Specifically, Section~\ref{sec2} states the general basics of VAD, including the task definition, type of anomalies, and application areas. Sections ~\ref{sec3}$\sim$\ref{sec5} elaborate on the learning paradigms and typical models of UVAD, WsVAD and FuVAD, respectively. We provide detailed explanations of classic methods for readers to understand the core ideas and implementations better. To assist readers to conduct research instantly, we provide a comprehensive introduction to existing datasets and evaluation metrics involved in the current work in Section~\ref{sec6}. Finally, Section~\ref{sec8} discusses the future vision of NSVAD, providing an in-depth analysis of its existing challenges, development trends, and possible opportunities. This tutorial paper is intended for non-specialists, so we prioritize conceptual clarification and research horizon construction in the main papers. Content that may overlap with existing work (e.g., itemized introductions to reviewed papers, detailed explanations of classical methods, and comprehensive presentations of research cases) is relegated to the \underline{\textit{Supplement}}.

\section{Generalized Foundations of NSVAD}~\label{sec2}

As a cross-cutting topic, NSVAD has attracted researchers from deep learning, video surveillance, mobile internet, and edge computing communities. Initially, VAD followed the conventional setting of AD problems, where anomalies were treated as outliers with different distributions \cite{ramachandra2020survey}. Corresponding unsupervised methods \cite{cao2024context,han2024mutuality,ye2024learning} aimed to learn prototypical representations of regular events, considering test videos outside the distribution as anomalies. To address video’s high-dimensional and complex backgrounds, AD researchers introduced efficient video representation learning techniques like Auto-Encoders (AEs) \cite{MNAD}, Generative Adversarial Networks (GANs) \cite{sun2024dual,huang2023selfsupervised,wu2024videobased}, Transformers \cite{CT-D2GAN}, Mamba \cite{li2024stnmamba}, and diffusion models \cite{basak2024diffusion,liu2025survey,li2023diffnas}. With large-scale video datasets and high-performance GPUs, deep learning-driven Video Understanding (VU) techniques have advanced, shifting VAD research to cross-cutting topics of VU and AD. New VAD routes emerged, such as WsVAD with multiple instance learning \cite{MIR} and FuVAD with iterative learning \cite{SDOR}. These methods challenge the open-world assumption under unsupervised AD, where real-world anomalies are varied and unbounded. WsVAD incorporates anomaly instances in training to differentiate between negative and positive samples. While WsVAD requires collecting and labeling anomalies, it outputs more reliable results for specific types of abnormal events. FuVAD avoids the data constraints of UVAD and WsVAD by learning anomaly detectors directly from raw videos, reducing data preparation costs and preventing mislabeling issues.

In recent years, the rise of multimodal learning \cite{xu2023multimodal,ektefaie2023multimodal} and large language models \cite{clip} have brought a new windfall for VAD. Researchers have proposed new tasks such as open-set VAD \cite{UBnormal,osvad}, open-vocabulary VAD \cite{ovvad}, and Video Anomaly Retrieval (VAR) \cite{wu2024toward}, indicating the trend of integration between VAD and generative AI research. We understand the cognitive differences resulting from the research backgrounds of these fields and view the emergence of new routes and tasks as a positive signal to promote VAD to systematic NSVAD research. In addition, NSVAD systems deployed in real-world scenarios must face the challenge of domain bias due to multi-view, cross-scenario videos and consider the limited storage and communication resources of end devices. Recent NSVAD advances \cite{amp,HL-Net} have begun considering both algorithm optimization and model deployment to balance performance and overhead. 

\subsection{Related Terms}~\label{sec21}

\begin{figure}
  \centering
  \includegraphics[width=.98\linewidth]{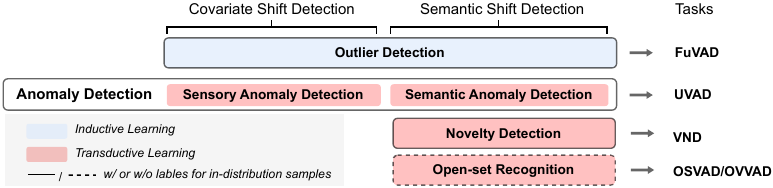}
  \caption{Illustration of AD-related terms' connection with NSVAD. The categorization is inspired by \cite{yang2021generalized}.}
  \label{tasks}
\end{figure}

We introduce key AD-related terms, including Anomaly Detection (AD), Novelty Detection (ND), Open Set Recognition (OSR), and Outlier Detection (OD). These terms often confuse researchers from the computer vision community \cite{wang2024survey,liu2024scmm}, and AD practitioners struggle with inconsistent definitions \cite{yang2021generalized}. We follow community consensus and our experience to clarify these terms, as shown in Fig.~\ref{tasks}.

\textbf{AD} detects samples that deviate from normality as defined by training data \cite{liu2022attentionbased}. Such deviations include: 1) covariate shift, i.e., label-independent distributional differences due to factors like image style or equipment, and 2) semantic shift, i.e., samples from different categories. The former is addressed by Sensory Anomaly Detection (SenAD), which includes tasks like domain adaptation. UVAD, by contrast, focuses on detecting anomalous samples with different semantic labels, termed Semantic AD (SemAD).

\textbf{ND} is often confused with AD since its goal is also to detect samples from unknown categories \cite{roady2020stream}. ND is modeled as a binary classification problem, identifying unknown categories without concern for secondary labels \cite{yang2021generalized}. Unlike SemAD, ND views unknown data positively, making UVAD similar to video novelty detection.

\textbf{OSR} trains a Multi-Class Classifier (MCC) to categorize in-distribution data while detecting unknown data during testing \cite{geng2020recent}. VAD generally does not categorize normal events but focuses on identifying anomalies. However, complex systems like autonomous cars require both anomaly detection and fine-grained categorization, leading to the integration of OSR and VAD in Open Vocabulary VAD (OVVAD).

\textbf{OD} detects outliers, samples significantly different from others \cite{boukerche2020outlier}. Unlike AD, ND, and OSR, which detect out-of-distribution samples only during testing, OD accepts all data types during training, similar to FuVAD. FuVAD handles unfiltered videos to learn anomaly classifiers, making it superior for large-scale real-time video streams in IoVT systems.

\subsection{Definition and Type of Anomaly}~\label{sec22}
Defining anomalies and understanding their types is essential for real-world NSVAD applications. UVAD and FuVAD follow the setups of SemAD and OD tasks, where anomalies are relative, meaning anything differing from common data is considered an anomaly \cite{cheng2023spatial}. In contrast, WsVAD focuses on specific pre-defined anomalies. 

Specifically, anomalies are usually categorized as sensory (raw data deviations) or semantic (label differences). VAD targets semantic anomalies, ignoring irrelevant factors like the scene or camera angle changes. Anomalies in UVAD and FuVAD fall into appearance-only, motion-only, or appearance-motion categories, corresponding to deviations in spatial, temporal, or spatial-temporal interactions. For example, in the CUHK Avenue dataset \cite{T3}, a red bag on a lawn is an appearance-only anomaly. More complex anomalies often involve misalignments in appearance-motion interactions, making them harder to detect with single-dimensional models. Therefore, effective UVAD models must understand regular event patterns in appearance, motion, and spatial-temporal contexts. Multi-proxy task-based models address this by improving the model's ability to distinguish between normal and anomalous events across different dimensions. WsVAD focuses on real-world hazardous events, such as crimes in the UCF-Crime dataset \cite{MIR} and violent incidents in XD-Violence \cite{HL-Net}. Although WsVAD cannot detect arbitrary anomalies, its results are more reliable. WsVAD anomalies are often categorized as short-term, long-term, or crowd anomalies, aligning with real-world concerns. 

\subsection{Definition of Various NSVAD Routes}

\begin{figure}
  \centering
  \includegraphics[width=\linewidth]{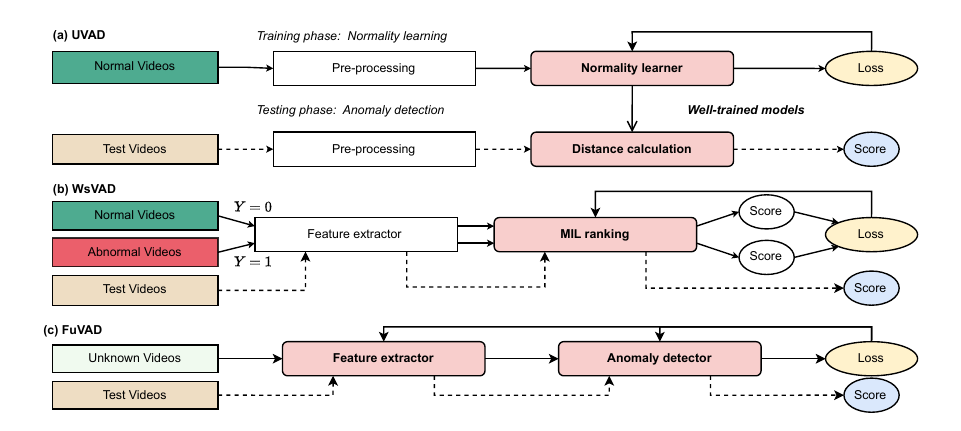}
  \caption{Illustration of general learning framework of (a) UVAD, (b) WsVAD, and (c) FuVAD research routes. 
 }
  \label{route}
\end{figure}

UVAD refers to NSVAD schemes that use only easily collected routine events to train models to learn the spatial-temporal pattern boundaries of normal samples \cite{ramachandra2020survey,zhang2024multi}. UVAD dominated early VAD research because it follows the open-world assumptions in the same vein as the AD community, circumventing predefinitions and collecting anomalous instances \cite{AST-AE,huang2024long}. In Fig.~\ref{route}, we show the general learning framework of UVAD and compare it with WsVAD and FuVAD. Specifically, UVAD assumes that models trained on regular events will only describe normal spatial-temporal patterns and will exhibit significant deviations when confronted with unseen anomalous examples, such as probability distributions \cite{T49,T4,T40,T52}, distances \cite{T68,LGA,T69}, and proxy task errors \cite{STM-AE,FF-AE,memAE,amp}. Early approaches first used local binary operators \cite{hu2018squirrel,nawarathna2014abnormal,zhang2015efficient}, spatial-temporal points of interest \cite{T72}, etc. to characterize spatial-temporal features that are normal events, and then employed One-Class (OC) classifiers (e.g., OC support vector machines and OC neural networks) \cite{OC-AE,sun2019abnormal} to learn the pattern boundaries, and considered test samples whose features fell outside the boundaries as anomalous. Such methods rely on manual features and are prone to dimensionality disasters. 

In recent years, deep learning-driven UVAD integrates feature extraction and normality learning into a unified framework with two phases, training and testing, which corresponds to normality learning via the use of negative samples and anomaly detection via the detection of out-of-distribution samples \cite{liu2024generalized}. In the normality learning phase, the network learnable parameters are optimized by minimizing a loss function overall negative samples. Whereas, in the testing phase, the degree of anomaly is measured by quantifying the distance between the test samples and the learned normality. Among them, reconstruction-based methods have dominated UVAD research in recent years \cite{FFP,memAE, MNAD,STM-AE, amp}. On the one hand, most of the challenges faced by such methods, such as global motion modeling, temporal normality learning, and spatial detail inference, have been intensively studied in video self-supervised learning. As a result, many methods have driven the development of UVAD by drawing inspiration from existing methods, such as video prediction \cite{vp,wang2022predrnn,zhao2025rethinking}. On the other hand, since reconstruction/prediction methods aim to learn a generative model that can reason about regular events, their basic settings and optimization goals are clear and unambiguous, and thus easy to implement and follow. Essentially, UVAD is transductive learning, i.e., model training and testing are relatively independent. Numerous studies have shown that the performance of models in the normality learning phase on the agent task does not show a positive correlation with downstream anomaly detection. Due to the diversity of events, the spatial-temporal features of normal and abnormal samples overlap, and UVAD usually fails to actively recognize discriminative features and incorrectly learns some shared patterns when only regular events are available for model training \cite{crc}. In addition, Park \textit{et al.} \cite{MNAD} pointed out that overpowered deep neural networks may be able to effectively reason about unseen anomalous events during the testing phase due to overgeneralization performance, which may lead to underdetection. In response, the researchers proposed a memory network enhancement approach to weaken the model's ability to generalize representations of anomalies by recording prototypical features.

WsVAD uses weakly semantic video-level labels to supervise the output of strong semantic frame-level labels, i.e., frame-by-frame anomaly scores, by the sequencing model, thus enabling temporal localization of anomalous events \cite{liu2022collaborative,karim2024real}. The first weakly-supervised approach is the multiple instance ranking framework proposed by Sultani \textit{et al.} \cite{MIR} in 2018, which lays out the basic MIL architecture of the WsVAD route, and whose concurrently publicly available UCF-Crime dataset has become the most widely used weakly-supervised benchmark. They consider a video as a collection of multiple examples (video clips), where clips containing abnormal frames are positive examples, while clips with all normal frames are labeled as negative examples. Obviously, a normal video with label 0 produces all examples called negative bags, and the example-level labels are all $0$. An abnormal video with the label $1$ constitutes a positive packet, which contains both positive and negative instances. Inspired by multiple instance learning, WsVAD aims to train a scoring model to output the anomaly scores of each example using video-level labels. The authors introduce a MIL ranking loss inspired by the hinge loss, which encourages the model to output high anomaly scores close to 1 for anomalous clips by maximizing the difference between the anomaly scores of the largest-scoring instances in the positive and negative bags while scoring regular clips as close to 0 as possible. In fact, WsVAD does not belong to any of the classes of out-of-distribution detection tasks introduced in Section~\ref{sec21}, but is rather a type of multiple instance learning under weak semantic labeling supervision. 

Compared to UVAD, the weakly supervised approach introduces anomalous videos in the training set and provides video-level labels for all training samples. Although anomalous examples are diverse and unenumerable in the real world, noteworthy anomalous events in specific scenarios are usually limited and easy to collect, such as thefts, robberies, and traffic accidents, which can be obtained in large quantities from surveillance IoT systems and online video platforms. In addition, the labor cost of video-level tagging, which only requires marking whether a video contains an anomaly without worrying about the specific timing location (frame-level labeling) and detailed anomaly category, is usually affordable. For example, the UCF-Crime dataset is much larger than the one used by UVAD, but only 1,900 discrete $[0,1]$ labels need to be provided. Due to the introduction of an additional human prior and the fact that the model has seen anomalous events during the training phase, WsVAD results are typically more reliable than UVAD, achieving excellent and consistent performance in detecting specific anomalous events. As a result, WsVAD has become a mainstream VAD scheme and is considered as the most promising research route for deployment in intelligent surveillance systems. The latest research attempts to mine anomaly-related clues from audio or subtitle text accompanying video frames, and proposes the multimodal WsVAD \cite{yang2024text}. 

FuVAD attempts to learn anomaly classifiers directly from large-scale raw videos without any editing and labeling \cite{SDOR, GCL}. Specifically, FuVAD's training data contains both positive and negative samples, and due to the low frequency of anomalous examples compared to regular events, the anomalous samples to be detected can be regarded as outliers with different patterns from the main data. In essence, the FuVAD model is transductive learning and does not follow the training-testing process of UVAD. Fully unsupervised methods have become a research hotspot in the internet era, where data preparation is costly, by virtue of the fact that they do not require any constraints on the training data, and can be used to train the model by directly accessing huge amounts of videos from the real world \cite{liu2024generalized}. 

\subsection{Emerging Research Tasks}~\label{sec24}

\begin{figure}
  \centering
  \includegraphics[width=0.8\linewidth]{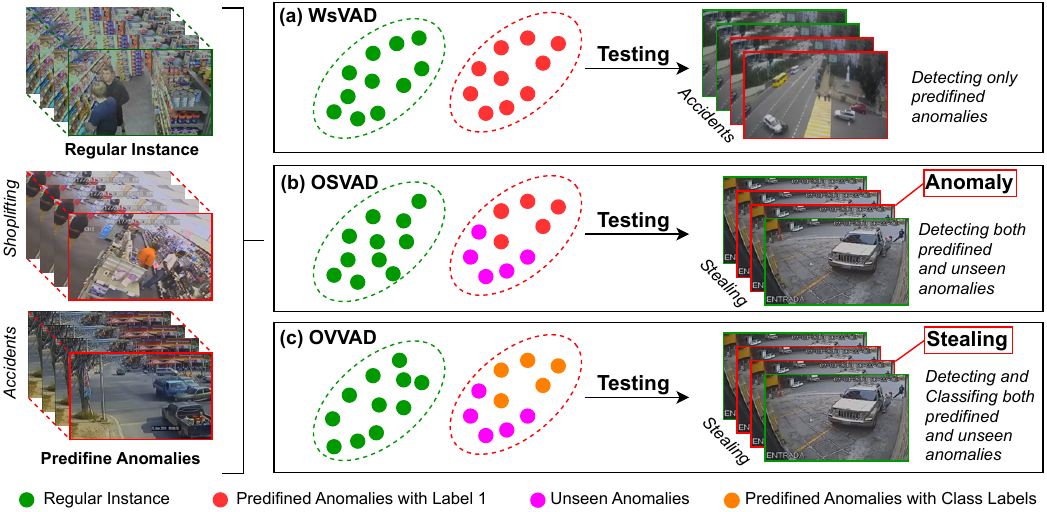}
  \caption{Illustration of the objectives of (a) WsVAD, (b) Open-Set VAD, and (c) Open vocabulary VAD. WsVAD only detects pre-defined types of anomalies in the training set, whereas OSVA has the open-set detection capability, which can recognize anomalies that have not been seen in the training phase. In contrast to (a) and (b), which treat anomalies as a single class, OVVAD can output specific semantic labels for both pre-defined and unseen anomalies.
 }
  \label{open}
\end{figure}

In Section~\ref{sec21}, we mentioned one of the inherent drawbacks of WsVAD, which violates the open-set problem property of the AD task by only detecting specific anomalies predefined in the training set and not being able to cope with diverse and arbitrary anomalous events in the open world. Action Recognition (AR) \cite{ar1,ar2} and VAD aim to understand specific behaviors. They can learn from each other's research ideas in data modality and feature learning, and the pre-trained AR models on large-scale video datasets are expected to be directly used for feature extraction and anomaly semantic cue mining in VAD. However, AR follows the closed-set task setting, which can only model the spatial-temporal patterns of known categories of behaviors but cannot empower the model to recognize out-of-distribution samples. In response, researchers have proposed various open-set VAD \cite{UBnormal,osvad} schemes to break through the above barriers, as shown in Fig.~\ref{open}(b). Zhang et al. \cite{zhang2024glancevad} introduced the concept of glance annotation, where a single frame from an anomalous event is randomly labeled and used as an enhanced supervision signal for training weakly supervised VAD models. They provided glance annotations for UCF-Crime and XD-Violence datasets, achieving a 5\% improvement in frame-level AUC compared to the state-of-the-art, demonstrating this setting’s outstanding potential for balancing annotation costs and model performance.

OpenVAD in \cite{osvad} aims to integrate the advantages of UVAD, which can handle arbitrary anomalous events, and WAED, which has a low false alarm rate in detecting specific anomalies. The proposed method integrates evidential deep learning and normalized flow into the MIL to equip WsVAD with the ability to identify unknown anomalies by quantifying uncertainty. Acsintoae \textit{et al.} \cite{UBnormal} propose a dataset for supervised OSVAD, named UBnormal, that maintains the task's open-set properties. Since the anomalous behavior in this dataset is generated through the VAD engine, it comes with fine pixel-level labels, making it possible to train VAD models in supervised learning. In short, this dataset attempts to bridge closed supervised learning and open anomaly detection, and experiments show that it can improve performance without compromising the open-set properties of existing VAD models. OVVAD  is closest in setting to the OSR task and aims to learn a multi-classifier capable of detecting and classifying all known and unknown anomalous events. Compared to OSVAD, OVVAD is more in line with the display requirements of scenarios such as autonomous driving. The first OVVAD model proposed by Wu \textit{et al.} \cite{ovvad} splits the task into two complementary tasks, i.e., AD and anomaly classification, and jointly optimizes them using the knowledge from the large models.

\subsection{VAD with Multimodal Large Language Models }~\label{sec25}
Large Language Models (LLMs) like Generative Pre-Training (GPT) \cite{radford2018improving} exhibit outstanding zero-shot learning and multimodal information processing abilities, showing great potential in VAD. Research has shown that multimodal LLMs can learn prototype patterns of normal events without training and describe any anomalies in open-set settings, significantly improving the generality and adaptability of VAD models. Zanella et al. \cite{zanella2024harnessing} proposed Language-based VAD (LAVAD), an unsupervised learning paradigm that leverages pre-trained LLMs and existing VLMs to train video anomaly detectors. They used VLMs to generate textual descriptions of video frames and designed a prompt mechanism to unlock LLMs' potential in temporal aggregation and anomaly score estimation, enabling direct VAD execution. Lv et al. \cite{lv2024video} introduced Video-LLaMA into VAD, aiming to break threshold limitations and improve model interpretability. The authors proposed a three-stage training method to improve the training efficiency of VLLM. In the AnomalyRuler, during the inductive stage, a small set of normal reference videos was provided to the LLM, enabling it to summarize normal patterns to induce rules for anomaly detection. In the Deduction stage, these induced rules were applied to detect anomalous frames in test videos. Hawk in \cite{tang2024hawk} uses an interactive VLM to accurately interpret video anomalies, answering VAD-related questions. It constructs an auxiliary consistency loss within the motion and video space, guiding the video branch to focus on motion modalities and establishing explicit supervision between actions and language to improve the accuracy of interpretation. Zhang et al. \cite{zhang2024holmes} developed a large-scale multimodal VAD instruction tuning benchmark called VAD-Instruct50k, used to build unbiased and interpretable VAD systems, as well as Holmes-VAD for anomaly event localization and interpretation.

\section{Unsupervised Video Anomaly Detection}~\label{sec3}

UVAD follows the general setup of semantic anomaly detection tasks, where only easily collectible regular instances are used to train models describing the normality of videos, detecting anomalies by measuring deviations between test samples and the learned model \cite{ijcai2024p77,cheng2024normality,wang2021stack}. From traditional machine learning to deep representation learning \cite{yang2024towards,wang2024confounded,yang2023spatio}, UVAD has undergone multiple advancements in feature extraction and normality learning, leading VAD to become a key issue in the AD and CV communities. With its close ties to the AD community and its long-standing development history, UVAD has long been regarded by researchers as the mainstream research route of NSVAD algorithms. As a result, existing surveys \cite{nayak2021comprehensive,ramachandra2020survey} typically focus on reviewing UVAD literature, lacking in-depth discussion of emerging weakly supervised \cite{MIST,RTFM} and fully unsupervised \cite{GCL,SDOR} routes, and overlooking novel tasks such as open-set and open-word detection. This article not only systematically reviews the latest developments in these new routes and tasks but also provides foundational knowledge and classic methods of UVAD in this section.

Based on the means of normality learning and deviation calculation principles, existing deep learning methods are generally divided into three categories: distance-based \cite{LGA,T68,T69}, probability-based \cite{T49,T4,T40,T52}, and reconstruction-based \cite{amp,liu2023stochastic,HSNBM,ConvLSTM-AE}. From our perspective, distance-based UVAD methods are a more general form of probability-based and reconstruction-based methods because probability deviation and reconstruction error calculation are fundamentally just different distance measurement methods. Specifically, distance-based methods include using single-classifier learning of video sequence spatial-temporal representations in the deep feature space, such as OC-SVM and OC-NN \cite{OC-AE,sun2019abnormal}. The drawback of such methods is that the trained models cannot be incrementally compatible with new data, leading to the need to retrain classifiers from scratch when new data is generated, such as in scenarios involving scene transitions. Another type of distance-based method is to use Gaussian mixture models \cite{T42} to model video feature normal vectors and measure deviations using Mahalanobis distance. In contrast, probability-based methods attempt to map the spatial-temporal representations of regular events into a probability framework and discriminate anomaly instances by measuring differences in probability distributions. Such methods tend to use traditional models such as Markov random fields \cite{T52} to build probability space. Deep learning-based attempts have encountered significant increases in computational costs and are noticeably slower in inference speeds.

In fact, the most prevailing deep UVAD approach is reconstruction-based \cite{ramachandra2020survey}. On the one hand, reconstruction-based methods \cite{FFP,MNAD,amp} aim to train models to represent the general spatial-temporal patterns of regular events through self-supervised proxy tasks, benefiting from advances in video self-supervised learning \cite{jing2020self} and deep neural networks. On the other hand, such methods avoid complex mathematical computations, which are easy to implement and exhibit excellent performance, thus being widely praised by existing researchers. The premise of reconstruction-based methods is that generative models trained on massive normal samples can effectively infer the spatial-temporal patterns of regular events. For anomalous events, the performance of proxy tasks will significantly decrease, and the resulting error can be used as a quantitative basis for measuring deviations to calculate anomaly scores. Common generative models include deep autoencoders \cite{FF-AE,T18,AMAE,STM-AE,MNAD}, variational autoencoders \cite{vae1,vae2}, and generative adversarial networks \cite{chen2021nm,nguyen2020anomaly,liu2025domain}. Proxy tasks include reconstructing input sequences and predicting future frames, which belong to pixel-level image generation. However, most reconstruction-based methods only calculate frame-level errors as anomaly scores without performing spatial localization, as the spatial contribution of anomalies is typically not significant. Spatial localization can only serve as a quantitative visualization result. 

\subsection{Taxonomy and Advances}~\label{sec31}

\begin{table}[t]
  \caption{Systematic Taxonomy of UVAD.}
  \label{taxo}
\begin{tikzpicture}[node distance=1cm,>=latex]
  \node (A) [rotate=90] {\textbf{UVAD}};
  \node (B) [right of=A, yshift=1cm, rotate=90] {GNL};
  \node (C) [right of=A, yshift=-1cm, rotate=90] {LPM};
  \node (D) [right=1cm of B,yshift=0.2cm,align=left,text width=6cm] {\begin{minipage}{10cm}
    \textbf{Single-task:} \cite{PC-LSTM}, \cite{FF-AE}, \cite{D-IncSFA}, \cite{DAF},\cite{ConvLSTM-AE},\cite{FFP},\cite{FFPN},\cite{LSTM-AE},\cite{STU-net}, \cite{AnomalyNet},\cite{ISTL},\cite{AnoPCN},\cite{memAE},\cite{MNAD},\cite{TSC},\cite{R-STAE},\cite{DD-GAN},\cite{AEP},\cite{STC-Net},\cite{Bi-Pre},\cite{ROADMAP},\cite{nayak2023video}
  \end{minipage}};
  \node (E) [right=1cm of B,yshift=-0.9cm,align=left,text width=8cm] {\begin{minipage}{10cm}
    \textbf{Multi-tasks:} \cite{CDD-AE},\cite{CDD-AE+},\cite{DSTAE},\cite{AMMC-net},\cite{AMAE},\cite{STM-AE},\cite{STAE},\cite{AMDN},\cite{AMC},\cite{GANs},\cite{OGNet}
  \end{minipage}};
  \node (F) [right=1cm of C,yshift=0.2cm, align=left,text width=6cm] {\begin{minipage}{10cm}
    \textbf{ST Patch:} \cite{ADCS},\cite{STCNN},\cite{DeepAnomaly},\cite{Deep-Cascade},\cite{DeepOC},\cite{ST-CaAE},\cite{AST-AE},\cite{S2-VAE},\cite{STC}
  \end{minipage}};
  \node (G) [right=1cm of C,yshift=-0.9cm,align=left,text width=6cm] {\begin{minipage}{10cm}
    \textbf{Object-level:} \cite{Background-Agnostic},\cite{Multi-task},\cite{LDGK},\cite{DCF},\cite{OC-AE},\cite{VEC},\cite{OAD},\cite{HF2VAD},\cite{BiP},\cite{HSNBM},\cite{osin}
  \end{minipage}};

  \draw [->] (A) -- (B);
  \draw [->] (A) -- (C);
  \draw [->] (B.south) -- (D.west);
  \draw [->] (B.south) -- (E.west);
  \draw [->] (C.south) -- (F.west);
  \draw [->] (C.south) -- (G.west);
  
\end{tikzpicture}
\end{table}

This article also systematically reviews deep learning-driven UVAD methods, providing a taxonomy that can aid in understanding the current state of research and inspire further exploration. However, due to space constraints and the emphasis of the tutorial paper on guiding beginners, we only present the underlying logic of the proposed taxonomy and summarize the research trends of the latest advances in the main text message. For a more detailed description of the existing UVAD methods, please refer to \underline{\textit{Section 1.1 of Supplement}}.

The methods based on distance and probability described earlier focus more on handcrafted features \cite{T40,T53} and traditional classification \cite{T100} models, which have been primarily surpassed by distance-based methods in the era of deep learning. Therefore, existing classification systems appear outdated in delineating recent advancements and reflecting research trends, failing to highlight the latest challenges and directions. To address this, inspired by data preprocessing techniques and forms of deep learning modeling, we categorize UVAD into two main classes: Global Normality Learning (GNL) and Local Prototype Modeling (LPM).

GNL utilizes the entire video sequence as input, often requiring no additional preprocessing such as spatial-temporal cube partitioning or foreground object extraction, and employs end-to-end deep neural networks to directly learn video ecologies \cite{graph3,memAE,MNAD,CDD-AE,amp,crc}. Over the years, researchers have believed that videos possess two informational dimensions, namely spatial and temporal, corresponding to appearance and motion, requiring different approaches to capture their normality. Therefore, within our UVAD taxonomy, GNL is further subdivided into Single-Proxy Task \cite{FFP,memAE,MNAD} and Multi-Proxy Task methods \cite{CDD-AE,STM-AE,amp}. In contrast, LPM methods argue that video data contain a plethora of redundant information tantamount to clues related to normality. Thus, they opt to use spatial-temporal cubes \cite{ADCS,STC,Deep-Cascade,AST-AE} or foreground objects \cite{HSNBM,osin} containing dense and effective information as network inputs instead of the entire video sequence, focusing on prototype feature learning of local patch. We classify such methods into spatial-temporal Patch-based and Object-driven methods.

UVAD methods can be grouped into single-task, multi-task, spatial-temporal patch-based, and foreground object-driven approaches. The first two categories, which input either full RGB frames or optical flow sequences, are considered global normality learning methods. In contrast, the latter two approaches focus on modeling local image patches or salient foreground objects, which are categorized under local prototype modeling. A taxonomy of these methods is presented in Table~\ref{taxo}, showcasing their latest developments and interrelations. Specifically, single-task methods treat the spatial and temporal patterns in video as entangled, typically employing a single network structure to execute a unified task for learning spatiotemporal normality. These methods are easy to design and train but may perform suboptimally in handling diverse anomalies in complex scenes. In comparison, multi-task methods regard appearance and motion as distinct information dimensions, using multi-branch networks (e.g., parallel autoencoders or encoder-decoder architectures) to perform different tasks for learning spatial and motion normality separately. This approach effectively handles anomalies involving appearance, motion, or a combination of both and has demonstrated outstanding performance in industrial, traffic, and medical applications. LPM methods address the redundant information in raw image sequences, which increases data processing costs and introduces noise that can degrade model performance. These methods first identify relevant spatiotemporal regions through preprocessing techniques before modeling them. Specifically, spatial-temporal patch (STP) methods assume that anomalies occupy small spatiotemporal regions in the video, and thus, modeling local cubes of data allows for precise spatiotemporal anomaly localization. Foreground object-driven (FOD) methods focus on analyzing patterns of foreground objects, leveraging pre-trained object detection models to extract the regions of interest for subsequent modeling.

\subsubsection{Global Normality Learning}~\label{sec321} Convolutional neural network-driven deep representation models can directly learn task-relevant spatial-temporal representations from raw video sequences and can adapt to different scale inputs and feature dimension requirements through simple structural adjustments. Global normality learning aims to learn video normality directly from complete RGB videos or optical flow sequences \cite{liu2023msn}. Compared to local prototype modeling, GNL requires no additional data preparation and is easy to optimize.

The earliest methods did not distinguish between spatial and temporal information, typically using only RGB videos as input and employing a single self-supervised proxy task (e.g.,  sequence reconstruction or future frame prediction \cite{ZHAO2024125581}) to broadly learn spatial-temporal normality. Generally, methods based on a single proxy task focus on designing more efficient single-stream end-to-end deep structures. Recent efforts include the introduction of 3D convolutional networks \cite{memAE} and convolutional long short-term memory networks \cite{ConvLSTM-AE} to enhance the representation capability of spatial-temporal features. Subsequent researchers discovered that spatial and temporal information have different characteristics, with the former focusing on local pixel inference and the latter on modeling global dynamics. Moreover, the addition of proxy tasks such as reconstruction and prediction losses often brings additional performance gains without significantly increasing training costs. Thus, they proposed introducing additional proxy tasks within the GNL framework. In addition to the separation of spatial-temporal normality learning \cite{STM-AE,CDD-AE+} and simple proxy task stacking \cite{STAE,Bi-Pre}, recent work has also explored novel tasks such as appearance-motion consistency \cite{AMMC-net}, spatial-temporal coherence \cite{cheng2023learning,ning2024memory}, and correlation \cite{STC-Net}.

\subsubsection{Local Prototype Modeling}

In contrast, local prototype modeling \cite{ADCS,STC,Deep-Cascade,AST-AE,HSNBM,osin} treats a video as an information cube with dimensions $h\times w \times l$, where $h$ and $w$ denote the spatial height and width, and $l$ represents the number of frames. It is observed that background information repeats across frames. On one hand, anomalies of interest typically occupy only a small portion of the information volume within the entire cube, and direct learning from the complete sequence often entails high computational costs. On the other hand, separating regions with different information densities and modeling their relationships with each other is beneficial for understanding the interaction between events. To address this, researchers have proposed the method of local prototype modeling, aiming to mitigate the handling of repetitive information to reduce training costs and model the relationship between foreground targets and background scenes to enhance anomaly detection performance. According to the data preprocessing methods, we categorize such methods into those based on Spatial-Temporal Patch-based (STP) and Foreground Object-Driven (FOD). The former employs simple spatial-temporal segmentation to divide the video into several information bodies, while the latter relies on pre-trained object detection models (e.g., RCNN \cite{RCNN}, FPN \cite{FPN}, and YOLO \cite{YOLO}) to selectively learn the spatial-temporal normality of specific subjects.

\subsection{Classic UVAD Models}~\label{sec33}

Considering the limited guiding value of a mere progress review for beginners, this article has chosen representative methods of UVAD to provide a detailed introduction to the research motivations and core ideas. The selected methods include: 1) Future Frame Prediction (FFP) framework \cite{FFP}, which introduces the video prediction into the VAD task for the first time. 2) Memory-Guided Normality Learning (MGNL) \cite{MNAD}, the first memory network for VAD. We elaborate on the implementation process and comprehensively review the related basic knowledge in \underline{\textit{Section 2.1 of Supplement}}.

\section{Weakly-supervised Video Anomaly Detection}~\label{sec4}

Inspired by multiple instance learning (MIL) \cite{mil}, WsVAD organizes videos into bags containing several instances. All segment instances from the same bag share a video-level label \cite{huang2022weakly}. Under this setting, $Y=0$ indicates negative bags while $Y=1$ indicates positive ones containing at least one anomalous instance. WsVAD strikes a balance between performance and data preparation cost, bridging the gap between UVAD with unreliable results and supervised learning with fine labels. Specifically, unlike unsupervised methods that train models only on regular events and simply consider all unseen samples as anomalies, WsVAD is trained with positive samples, enabling it to effectively understand the inherent differences between normal and anomalous instances. Consequently, existing research indicates that WsVAD yields more reliable results and outperforms UVAD schemes with lower false alarm rates in real-world tasks, such as crime behavior identification and traffic accident detection \cite{liu2023distributional}. Annotating long video sequences frame by frame for supervised learning is often impractical. For instance, the training set of the UCF-Crime dataset includes 1,610 long videos containing 13,741,393 frames in total. In contrast, WsVAD only requires coarse semantic video-level annotations, with each video needing only a discrete binary label. Various types of anomalies are roughly labeled as 1, requiring minimal expert involvement in data annotation and significantly reducing labor costs compared to fine-grained labeling. Thus, although WsVAD no longer adheres to the open-world task setting of the AD community and can only detect specific anomalies, it has garnered widespread attention in recent years due to its stable performance and reliable results. Subsequent works typically adopt the MIL regression task setting proposed by Sultani \textit{et al.} \cite{MIR} and utilize their concurrently open-sourced UCF-Crime dataset as a benchmark. To further validate the generalization ability of weakly-supervised models, some researchers relocate positive samples from the test set of UVAD datasets to the training sets and provide video-level labels, proposing reconfigured datasets such as UCSD Ped2 and ShanghaiTech Weakly for WsVAD validation.

\subsection{Taxonomy and Advances}

The superior performance of WsVAD on real-world videos has inspired researchers to develop VAD models tailored to the complexities of real-life scenarios. In 2020, Wu \textit{et al.} \cite{HL-Net} extended the research scope of VAD from single-modal video pattern analysis to multimodal learning to leverage heterogeneous data in real-world scenes for enhanced anomaly event detection. Their collected XD-violence dataset is the first multimodal VAD benchmark, comprising RGB images and audio modalities, focusing on violence behavior detection in complex scenes. This work not only provides the first multimodal VAD dataset and solution but also motivates researchers in the community to explore anomaly clues from multimodal data such as audio and text, leading to a new wave of VAD development: multimodal WsVAD. In this section, we comprehensively review the latest advancements in both single-modal and multimodal WsVAD, and illustrate the foundational knowledge and specific implementations required for WsVAD using the Multiple Instance Regression (MIR) framework and the Local-Global Network (HL-Net) as examples.

Unimodal methods take only visual data as input and attempt to learn the pattern differences between normal and anomalous events based on the appearance and motion semantics reflected in the RGB sequences. In contrast, multimodal methods explore additional data modalities beyond visuals, such as audio and text, using them as complementary semantics to improve the model’s anomaly detection capabilities. Although the inclusion of additional modalities increases the data processing and model training costs, existing studies show that audio and text can significantly enhance WsVAD’s ability to detect anomalies, especially in cases where visual information alone cannot distinguish anomalies effectively. Since multimodal methods still need to model spatiotemporal patterns in the visual modality and continue to use the same weak supervision task setup and multi-instance learning strategies as unimodal approaches, certain strategies already explored and validated in unimodal WsVAD—such as dataset bias correction, label noise reduction, and hinge loss optimization—can be incorporated into multimodal WsVAD methods, particularly for the visual data processing branch.

\subsubsection{Unimodal Methods}
Unimodal WsVAD (UWsVAD) methods focus on extracting anomaly-related cues from RGB image sequences \cite{STA}. These methods generally follow a three-step process: 1) preprocessing videos into several non-overlapping segments, 2) extracting spatial-temporal features using models like Convolutional 3D Networks \cite{C3D} and Inflated 3D Networks \cite{I3D}, and 3) computing anomaly scores using a multi-instance ranking loss to differentiate between normal and anomalous instances. The Multi-Instance Ranking (MIR) framework \cite{MIR} first introduced MIL to WsVAD with a focus on predicting higher anomaly scores for anomalous segments while minimizing score fluctuations. For a more detailed description of the existing UWsVAD methods, please refer to \underline{\textit{Section 1.2.1 of Supplement}}. Moreover, given that VsVAD datasets are typically collected from real-world scenarios and contain identity-sensitive information such as faces, Fioresi et al. \cite{fioresi2023ted} proposed a privacy-preserving VAD framework named TeD-SPAD. They first anonymized video frames using a UNet to eliminate privacy information before using the I3D network to extract spatial-temporal features. Results showed that TeD-SPAD successfully prevented 32\% of visual information leakage.

\subsubsection{Multimodal Methods}
Multimodal approaches in WsVAD integrate various data types, primarily focusing on the fusion of video with audio \cite{ACF} and text \cite{wu2024toward}. These methods face significant challenges due to limited benchmark datasets and standard comparison metrics, which have hindered their widespread adoption. 

One of the hallmark advancements in multimodal VAD is the effective fusion of video and audio data. Existing research \cite{MACIL-SD,AGAN,MSAF} primarily uses the XD-Violence \cite{HL-Net} dataset as the evaluation benchmark, which introduces the audio into video violence detection. More information about the video-audio-based VAD methods is provided in \underline{\textit{Section 1.2.2 of Supplement}}.

Moreover, with the rise of Visual-Language Learning (VLL), particularly through the emergence of LLMs, researchers have begun to leverage textual information to further enhance VAD models' performance. Pre-trained Vision Vision LLMs can describe appearance and motion information in videos without prior samples, embedding such text as prompts into visual representations. This integration represents a significant advancement in the fusion of video and text, which improves the model's ability to express complex anomalies. The incorporation of textual information not only enhances generalizability and interpretability but also gives rise to new research tasks of practical value, such as video anomaly retrieval \cite{wu2024toward} and open vocabulary VAD \cite{ovvad}, as illustrated in Sections \ref{sec24}. 

Specifically, VLL models \cite{yang2024stephanie} can provide accurate textual descriptions of video frames, enhancing the semantic mining capability of existing vision-based VAD models. For instance, Chen et al. \cite{chen2023tevad} used a language model-based captioning network to obtain textual descriptions of video sequences, which, after being embedded in a text embedding network, were fused with visual features as inputs to the anomaly detector. Their proposed Text Empowered Video Anomaly Detection (TEVAD) efficiently captures abstract semantics of anomaly events and enhances the interpretability of VAD models. Pu et al. \cite{pu2024learning} introduced a Prompt-Enhanced Learning (PEL) module, using knowledge-based prompts to incorporate semantic priors, improving the discriminative power of visual features in weakly supervised VAD, while ensuring separability between anomalous subcategories. Wu et al. \cite{wu2024toward} proposed VAR, which efficiently detects specific anomalies based on cross-modal learning (e.g., language descriptions and synchronized audio). They designed an Anomaly-Led Alignment Network (ALAN) using BERT \cite{kenton2019bert} to process text information and incorporate a pretext task to enhance semantic alignment between video-text fine-grained representations.

In addition, visual-language associations can serve as effective cues for detecting video anomalies, and the pre-learned visual-text consistency in large VLL models can be efficiently transferred to VAD. Kim et al. \cite{kim2023unsupervised} used large language models to generate textual descriptions of video frames and detected anomalous frames by calculating the cosine similarity between input frames and their textual descriptions using CLIP. Text Prompt with Normality Guidance \cite{yang2024text} leverages the language-visual knowledge of the CLIP model to align video frames with textual descriptions of events, generating more accurate pseudo-labels for WsVAD, thus improving model performance. 

In surveillance videos, it is often difficult to capture synchronized audio or text data, making unimodal methods the mainstream approach. However, with the rise of live streaming platforms and the film industry, multimodal methods will play a crucial role in online video moderation and content detection for TV shows, films, and animations. On the one hand, the context of such content is often highly varied, with anomalies taking many forms, such as visual violence or non-compliant audio or text, making it difficult for unimodal methods relying solely on RGB sequences to handle effectively. On the other hand, these types of videos typically capture synchronized audio and provide subtitles or other multimodal data, allowing multimodal methods to function without additional data preparation costs.

\subsection{Classic WsVAD Models}

We select two representative methods from unimodal and multimodal WsVAD, i.e.,  MIR \cite{MIR} and HL-Net \cite{HL-Net}, to elaborate on the concepts of MIL and multimodal information processing in WsVAD research, respectively. Specifically, MIR has laid the foundation for the MIL-based solution, marking a milestone in VAD research. However, its optimization objectives and detailed implementation are challenging for many researchers in the AD and CV communities to understand. Therefore, we provide a comprehensive exposition of the motivation, theoretical logic, and related knowledge. In contrast, HL-Net is the first model for multimodal violence detection. The simultaneously released XD-violence dataset has inspired lots of researchers from the AD and multimodal understanding fields to delve into the emerging hotspot of multimodal VAD. We choose this method as a case study to disseminate knowledge of multimodal understanding to the AD community, aiming to propel multimodal VAD from simple-modal fusion towards systematic multimodal anomaly clues exploration, please see \underline{\textit{Section 2.2 of Supplement}} for more information.

\section{Fully unsupervised video anomaly detection}~\label{sec5}

FuVAD follows the transductive learning setup of traditional outlier detection tasks, aiming to directly learn an anomaly classifier from all unfiltered observations to detect samples significantly different from the primary data. Existing methods, partly inspired by time series outlier detection research, utilize deep clustering to identify pattern centers of the data and consider samples far from the learned centers as anomalies. However, the pattern dimension of video data is much higher than that of time series, and due to reasons such as similar environmental backgrounds, normal and abnormal event spatial-temporal patterns often overlap, making it infeasible to determine pattern boundaries through clustering when dealing with complex datasets. In recent years, researchers have proposed FuVAD schemes based on iterative learning, gradually amplifying the pattern differences between anomalous samples and dominant normal data through the cooperation of feature extraction modules and anomaly models. Compared to other VAD research routes, FuVAD does not require filtering and labeling training data, directly utilizing unclipped unlabeled monitoring videos to train models, which aligns with the data state-agnostic condition in online learning.

\subsection{Recent Advances}

Inspired by unmasking, Liu \textit{et al.} \cite{HS} connect heuristic unmasking with multiple classifier two-sample tests, introducing a history sampling method to enhance testing capabilities in video anomaly detection and a motion feature calculation method for better representation and generalization. Li \textit{et al.} \cite{Deep-UAD} use a distribution clustering to identify anomaly example groups, then train an autoencoder with normal data subsets to learn representations of normalcy, iterating this process to refine the encoder's ability to describe regular events.

Drawing from the Masked Autoencoder (MAE) \cite{he2022masked}, the Temporal Masked Auto-Encoder (TMAE) \cite{TMAE} aims to learn high-quality representations for anomaly detection by employing a visual transformer for completion tasks on spatial-temporal cubes, recognizing the significance of the temporal dimension in video anomalies. This approach is designed to efficiently complete regular events, highlighting anomalies due to their significant loss during completion.

An end-to-end self-training deep ordinal regression (SDOR) framework \cite{SDOR} iteratively learns pseudo-normal and anomaly scores from raw sequences, starting with identifying potential anomaly frames using existing algorithms and employing ResNet50 and neural networks for score computation, leveraging self-training for simultaneous optimization of feature learning and anomaly scoring. Generative Cooperative Learning (GCL) \cite{GCL} learns anomaly detectors from unlabeled mixed data by exploiting anomaly events' low frequency, featuring a generator $\mathcal{G}$ and discriminator $\mathcal{D}$ working cooperatively. The $\mathcal{G}$ focuses on regular event representations and uses negative learning for anomalies, generating pseudo-labels for $\mathcal{D}$, which estimates anomaly probabilities to further refine $\mathcal{G}$.

\subsection{Classic FuVAD Models}

We introduce two typical deep learning-based FuVAD schemes in \underline{\textit{Section 2.3 of Supplement}}, including 1) Self-trained Deep Ordinal Regression (SDOR) \cite{SDOR}, which utilizes self-training to jointly optimize feature learning and anomaly scorer, and 2) Generative Cooperative Learning \cite{GCL}, which leverages the low-frequency nature of real-world anomalies to construct pseudo-labels for FuVAD. SDOR \cite{SDOR} is the first deep NSVAD method designed for unfiltered and unlabeled videos, which explicitly points out the limited applicability of UVAD and WsVAD in realist scenarios due to the constraints on data and the performance reliance on feature learning. GCL \cite{GCL} introduces negative learning to increase the contrast between regular sequences and potential anomalies, directly learning the differences between anomalies and the majority of samples (normal) through the interplay of generator and discriminator.

\section{Model Evaluation}~\label{sec6}

\begin{table}[t]
  \centering
  \caption{Statistical results of the NSVAD dataset.}
  \label{tab:dataset}
  \begin{threeparttable}
  \resizebox{\textwidth}{!}{
  \begin{tabular}{@{}cccccc|cccccccc@{}}
  \toprule
  \multirow{2}{*}{\textbf{Labeling}} & \multirow{2}{*}{\textbf{Year}} & \multirow{2}{*}{\textbf{Dataset}} & \multicolumn{3}{c}{\textbf{\#Videos}} & \multicolumn{5}{c}{\textbf{\#Frames}} & \multirow{2}{*}{\textbf{\#Scenes}} & \multirow{2}{*}{\textbf{\#Classes}} & \multirow{2}{*}{\textbf{\#Anomalies}} \\ \cmidrule(lr){4-11}
 & &  & Total& Training & Testing & Total & Training & Testing & Normal  & Abnormal && & \\ \midrule
  \multirow{9}{*}{Unsupervised}  & 2008  & \href{https://vision.eecs.yorku.ca/research/anomalous-behaviour-data/sets/}{Subway Entrance} & - & -  & - & 144,250& 76,543 & 67,797& 132,138  & 12,112 & 1 & 5  & 51\\
 & 2008  & \href{https://vision.eecs.yorku.ca/research/anomalous-behaviour-data/sets/}{Subway Exit}  & - & -  & - & 64,901  & 22,500& 42,401  & 60,410  & 4,491 & 1 & 3  & 14\\
 & 2011  & \href{http://mha.cs.umn.edu/proj_events.shtml\#crowd}{UMN}$^\dagger$ & - & -  & - & 7,741  & -  & - & 6,165 & 1,576  & 3 & 3  & 11\\
 & 2013  & \href{http://www.svcl.ucsd.edu/projects/anomaly/dataset.htm}{UCSD Ped1}& 70 & 34  & 36 & 14,000 & 6,800  & 7,200 & 9,995 & 4,005  & 1 & 5  & 61\\
 & 2013  & \href{http://www.svcl.ucsd.edu/projects/anomaly/dataset.htm}{UCSD Ped2} & 28 & 16  & 12 & 4,560  & 2,550  & 2,010 & 2,924 & 1,636 & 1 & 5  & 21\\
 & 2013  & \href{http://www.cse.cuhk.edu.hk/leojia/projects/detectabnormal/dataset.html}{CUHK Avenue}  & 37 & 16  & 21 & 30,652 & 15,328 & 15,324& 26,832& 3,820  & 1 & 5  & 77\\
 & 2018  &  \href{https://svip-lab.GitHub.io/dataset/campus_dataset.html}{ShanghaiTech} & - & -  & - & 317,398& 274,515& 42,883& 300,308  & 17,090 & 13& 11 & 158  \\
 & 2020  & \href{https://www.merl.com/demos/video-anomaly-detection}{Street Scene} & 81 & 46  & 35 & 203,257  & 56,847& 146,410 & 159,341 & 43,916 & 205 & 17 & 17\\
 & 2023  & \href{https://campusvad.GitHub.io/}{NWPU Campus}  & 547& 305 & 242& 1,466,073  & 1,082,014  & 384,059  & 1,400,807 & 65,266 & 43& 28 & -\\ \hline
  \multirow{3}{*}{\begin{tabular}[c]{@{}c@{}}Weakly\\ Supervised\end{tabular}} & 2018  &  \href{https://webpages.charlotte.edu/cchen62/dataset.html}{UCF-Crime} & 1,900 & 1,610& 290& 13,741,393 & 12,631,211 & 1,110,182 & - & -  & -& 13 & 950  \\
 & 2019  &  \href{https://github.com/jx-zhong-for-academic-purpose/GCN-Anomaly-Detection/}{ShanghaiTech Weakly}  & 437& 330 & 107& -  & -  & - & - & -  & -& 11 & -\\
 & 2020  & \href{https://roc-ng.GitHub.io/XD-Violence/}{XD-Violance}  & 4,754 & 3,954   & 800 & -  & -  & - & - & -  & -& 6  & -\\
 & 2020  & \href{https://github.com/ktr-hubrt/WSAL}{TAD}  & 500 & 400  & 100 & 540,272  & -  & - & - & -  & -& 7  & 250\\
 \hline
 Supervised & 2022  & \href{https://github.com/lilygeorgescu/UBnormal}{Ubnormal}$\ddagger$& 543& 268 & 211& 236,902& 116,087& 92,640& 147,887  & 89,015 & 29& - & 660  \\ \bottomrule
  \end{tabular}}
  \begin{tablenotes}    
    \footnotesize 
    \item $^\dagger$The frame rate is set to 15 fps. $\ddagger$The Ubnormal contains a validation set with 64 videos totaling 14,237 normal and 13,938 abnormal frames.
  \end{tablenotes}   
\end{threeparttable}
  \end{table}

\begin{figure}
  \centering
  \includegraphics[width=\linewidth]{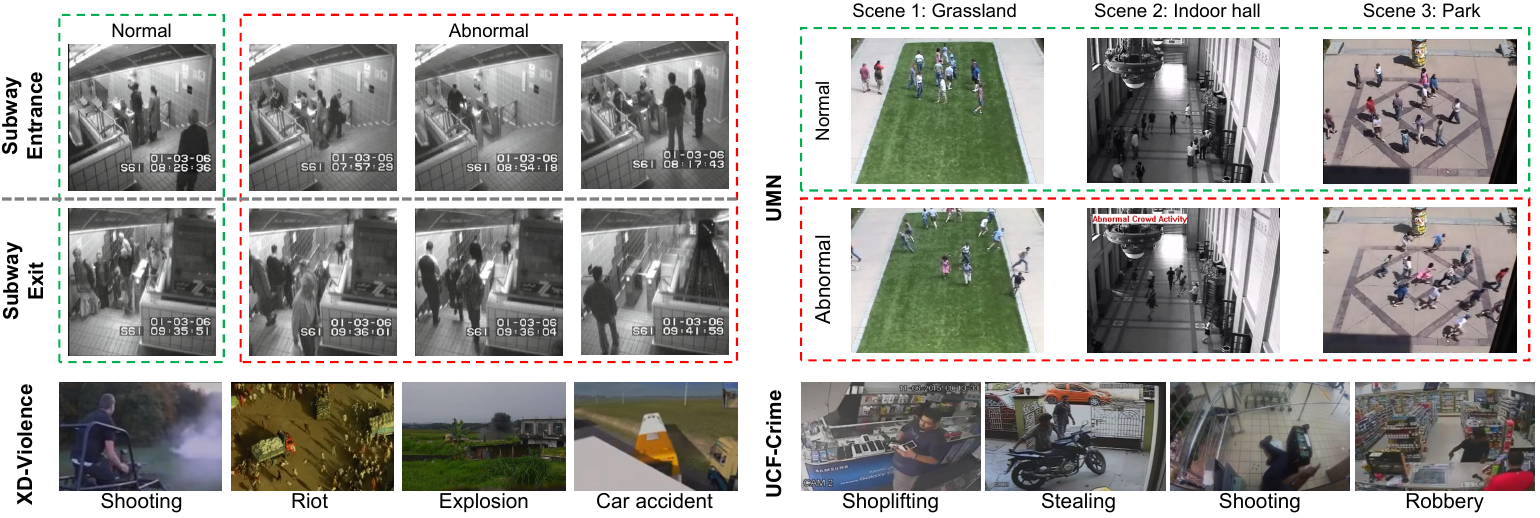}
  \caption{Exapmles of classical UVAD (Subway \cite{Subway} and UMN \cite{UMN}) and WsVAD (UCF-Crime \cite{MIR} and XD-Violence \cite{HL-Net}) dataets.}
  \label{subway}
\end{figure}

In this section, we elaborate on the characteristics of prevailing datasets and common evaluation metrics. Based on the annotations, we categorize existing datasets into unsupervised, weakly supervised, and supervised, as presented in Table~\ref{tab:dataset}. The examples of UVAD (e.g., Subway \cite{Subway} and UMN \cite{UMN}) and WsVAD (e.g., UCF-Crime \cite{MIR} and XD-Violence \cite{HL-Net}) datasets are illustrated in Fig.~\ref{subway}. Since such datasets have been extensively surveyed, we only present their details in \underline{\textit{Section 4 of Supplement}}.
We provide a systematic WSVAD evaluation system in this section, categorizing existing metrics into accuracy-oriented and cost-oriented as well as introducing system-level performance indicators.

\subsection{Conventional Evaluation}

\begin{figure}
  \centering
  \includegraphics[width=\linewidth]{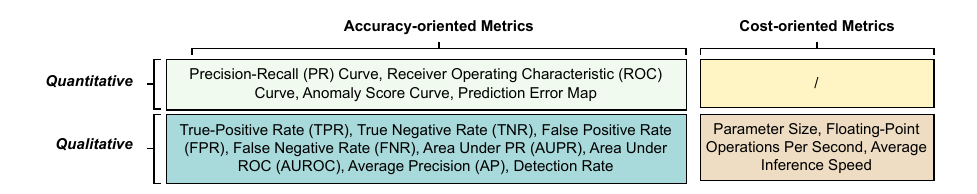}
  \caption{Conventional evaluation metrics. 
 }
  \label{metric}
\end{figure}

The existing works typically evaluate the proposed method from two perspectives: detection accuracy and model cost. On the one hand, models are expected to detect noteworthy anomalies as accurately as possible. Considering the varying influences of false positives and false negatives, along with the highly imbalanced data, researchers have proposed multiple quantitative metrics such as Area Under the Receiver Operating Characteristic curve (AUROC), Area Under Precision-Recall curve (AUPR), and detection rate, to assess model accuracy. Additionally, anomaly score curves are commonly used to qualitatively demonstrate the sensitivity to abnormal intervals, while prediction error maps are widely employed in reconstruction-based UVAD methods to visualize the performance of spatial localization. On the other hand, metrics like parameter size and inference speed determine whether the model can be deployed on resource-constrained devices. Thus, we categorize conventional metrics based on the orientation and evaluation dimension, as illustrated in Fig.~\ref{metric}.

\subsubsection{Accuracy-oriented metrics} 
Accuracy-oriented metrics aim to evaluate a model's ability to distinguish between normal and abnormal events, including quantitative metrics such as AUROC, AUPR, false alarm rate, and detection rate, as well as qualitative metrics like anomaly score curves and prediction error maps. While the task definitions vary across different VAD routes, they all aim to learn an anomaly detector capable of quantitatively measuring the abnormality level of test samples. Specifically, UVAD is a one-class classification task to train a model using regular events to describe normal patterns while considering all uncharacterizable samples as anomalies. The abnormality degree is computed by measuring the deviation to the learned normality model, which is typically normalized to the range of $[0,1]$ as anomaly scores. In contrast, WsVAD treats VAD as a regression task, using video-level labels to supervise fully connected networks directly outputting instance-level anomaly scores, similar to the FuVAD model. Therefore, despite differences in task settings and anomaly discrimination processes, metrics from binary classification tasks can be used to evaluate VAD models.

In most cases, the anomaly scores computed by NSVAD models are continuous values in the range $[0,1]$, while the given data labels are binary discrete values, where 0 denotes negative (normal events) and 1 represents positive (anomalous events). Therefore, it's necessary to select a threshold to convert relative abnormality scores into definitive binary labels for comparison. For example, with a threshold of 0.5, samples with scores lower than 0.5 are considered negative by the model, while those greater than or equal to 0.5 are considered positive. Thus, we can compute True Positive Rate ($TPR$), False Positive Rate ($FPR$), True Negative Rate ($TNR$), and False Negative Rate ($TNR$), as follows:
\begin{equation}
 TPR = \frac{TP}{TP+FN}, FPR = \frac{FP}{FP+TN}, TNR = \frac{TN}{FP+TN}, FNR = \frac{FN}{TP+FN}
\end{equation}
where $TP$, $FP$, $FN$, and $TN$ represent correctly detected positive samples, negative samples misclassified as positive, correctly detected negative samples, and positive samples misclassified as negative, respectively. 

Due to the highly imbalanced nature of positive and negative samples in NSVAD, some common evaluation metrics for classification tasks, such as accuracy, are not applicable. For example, a model biased towards outputting label 0 thus missing anomalous events, would be incorrectly evaluated as good under such metrics. Using a single threshold to simply assess a model's ability to differentiate between normal and abnormal patterns is not wise. For instance, with a threshold of 0.5, a model that consistently outputs scores slightly below 0.5 for regular events and slightly above 0.5 for anomalies would be considered perfect because it would show optimal performance across various metrics. However, such a model may not have learned the inherent differences between normal and abnormal patterns well, resulting in minimal score gaps between the two, which could lead to failure in detecting subtle anomalies and normal instances with data bias in complex scenarios. Therefore, researchers have introduced Receiver Operating Characteristic (ROC) curves, which measure VAD models more comprehensively by selecting multiple thresholds. Specifically, this curve plots the TNR against the TPR at various thresholds. The area under the  curve, known as AUROC, has been the most widely used VAD evaluation metric. An ideal AUROC value of 1 indicates a model that outputs a score of 0 for all negative samples and 1 for all positive samples, aligning with our expectations. Considering that TN usually exceeds TP, researchers argue that Average Precision, i.e., the Area Under the Precision-Recall (AUPR) curve, is more suitable for evaluating anomaly detection tasks. The PR curve depicts precision and recall (i.e., TPR) at specific thresholds. The point on this curve where Precision equals Recall is the balance point. Currently, multimodal anomaly detection models primarily use AP for quantitative evaluation. An anomaly score curve is commonly used to intuitively demonstrate the model's response to anomalous events, presenting the temporal localization capability. In contrast, prediction error maps are often used to assess UVAD model's spatial localization capability.

\subsubsection{Cost-oriented metrics}
Current research primarily focuses on developing high-performance detection models while neglecting to evaluate the models' deployment potential. A lightweight model is crucial in driving the application of NSVAD. Thus, we compile deployment-oriented metrics, including parameter size, Floating-Point Operations Per Second (FLOPS), and average inference speed. Specifically, parameter size indicates the number of learnable parameters, reflecting the complexity and storage cost of the model. In the real world, while complex models may offer performance gains, the resulting increase in memory and computational resource requirements may be unacceptable. Therefore,  NSAVD should balance detection performance with model parameters. FLOPs represent the number of floating-point operations the model needs to perform during inference. This metric is crucial for end devices, as excessive FLOPs may lead to performance bottlenecks and reduced hardware lifespan. Some existing works report average inference speed, i.e., the number of frames the model can process during testing, to quantitatively reflect the model's run time. However, due to inconsistent experimental environments, it cannot serve as an instructive and convincing metric in most cases. Due to space constraints and fairness of the comparison, we have collected the performance reported by existing methods but only organized these results in our GitHub\footnotemark[1] repository for reference.

\subsection{System-level Evaluation}

While conventional metrics focus on the detection accuracy and cost of individual models, evaluating the performance of NSVAD in real-world deployments requires a more holistic, system-level perspective. Beyond model-specific metrics, the entire system's effectiveness depends on various factors such as latency, communication cost, bandwidth efficiency, data security, user privacy, and system robustness \cite{liu2024privacy}. These metrics collectively assess the performance and feasibility of deploying NSVAD in large-scale, distributed, and resource-constrained environments. To this end, we categorize system-level metrics into three primary groups: efficiency, privacy, and robustness. 
\subsubsection{Efficiency-oriented metrics}
Efficiency-oriented system metrics assess the efficiency and scalability of the NSVAD system in a distributed setting. One key metric is {latency}, which measures the end-to-end delay from the moment video data is captured to the final detection output. For real-time anomaly detection, minimizing latency is critical, particularly in scenarios such as public safety monitoring or autonomous driving, where any detection delay could result in catastrophic outcomes. Latency can be broken down into communication latency, processing latency, and system response time, each reflecting different aspects of delay within the system. Reducing these latencies often requires optimizing the placement of inference tasks across edge and cloud nodes.
{Communication cost} evaluates the data transmission overhead between distributed nodes, especially in edge-cloud architectures. Given the high bandwidth demands of video data, optimizing communication efficiency becomes crucial for deploying NSVAD at scale. Common metrics used here include the total data transmitted (measured in megabytes or gigabytes) and the number of communication rounds required for model updates in federated learning-based systems. To address these challenges, techniques such as video compression, parameter pruning, and model quantization are often employed to reduce communication overhead.
Another essential metric is {bandwidth utilization}, which measures the efficiency of the network resources. High bandwidth usage may congest the network, causing delays and performance degradation, particularly in multi-client or large-scale environments. Methods such as asynchronous communication and bandwidth allocation prioritization can be leveraged to optimize utilization without sacrificing detection performance.

\subsubsection{Privacy-oriented metrics}
In NSVAD systems, maintaining data security and user privacy is paramount, particularly in applications involving sensitive environments such as healthcare \cite{li2024silent} or public surveillance. {Data security} is typically evaluated using metrics such as encryption overhead, which measures the computational cost introduced by encryption techniques, and key management efficiency, which assesses how well the system handles the distribution and renewal of cryptographic keys in a large-scale deployment. Robust encryption algorithms, such as AES or homomorphic encryption, are commonly used to ensure that video data remains secure during transmission and processing.
{User privacy} is another critical concern in video anomaly detection systems. Metrics such as the {privacy leakage rate} evaluate how much sensitive information (e.g., identities, personal activities) can be inferred from the system’s outputs. Differential privacy, federated learning, and encrypted video coding are among the techniques employed to minimize privacy risks. Privacy-preserving methods are evaluated based on the degree of anonymization they provide, often measured in terms of the {privacy budget}, which balances privacy protection against utility loss.

\subsubsection{Robustness-oriented metrics}
Robustness metrics aim to assess the system's ability to maintain reliable performance in the presence of adversarial conditions, such as network disruptions, data corruption, or malicious attacks. One critical metric is {fault tolerance}, which measures the system's capacity to continue operating when certain components fail \cite{guo2025adaptive,guo2025feature}. This is especially important in distributed settings where failures in edge devices or communication links can affect the overall detection pipeline. Techniques such as redundancy, dynamic task migration, and edge-cloud coordination can enhance the system’s fault tolerance.
{Adversarial robustness} evaluates the system's resilience to attacks designed to manipulate or mislead the anomaly detection model. Adversarial attacks may involve injecting malicious data, such as perturbing video frames or manipulating model parameters. The robustness against such attacks is typically quantified by the system’s ability to maintain high detection accuracy even when exposed to adversarial perturbations. Metrics such as adversarial success rate and robust accuracy are often used.
Lastly, {scalability} measures the system’s ability to handle increasing workloads, including the number of video streams and distributed clients. This is typically evaluated through stress testing, where system performance is analyzed under different load conditions to ensure it can maintain efficiency and reliability as deployment scales up. A scalable NSVAD system should efficiently distribute workloads across edge devices and cloud servers without degrading performance.

\section{Discussion and Summary}~\label{sec8}
\subsection{Research Challenges}

Sections~\ref{sec3}-\ref{sec5} introduced the key challenges addressed by various NSVAD algorithms. Here, we further elaborate on unresolved problems from the perspectives of data, labels, models, and systems. In contrast to previous works that emphasize only algorithm design, we also explore the NSVAD-specialized bottlenecks encountered in real-world deployments, such as communication and computing overhead, large-scale detection demand, and privacy concerns.

\subsubsection{Data}
Real-world videos exhibit label-independent domain shifts due to variations in scenes, camera angles, and device configurations \cite{duong2023deep}. These subtle differences in spatial-temporal patterns, while easily comprehended by humans, often lead to high false positive rates in NSVAD models \cite{crc}. Existing methods typically validate their models on datasets from a single scene to avoid this issue \cite{ramachandra2020survey}. For instance, the UCSD dataset \cite{T2} includes two distinct perspectives, but they are treated as separate datasets. The ShanghaiTech dataset \cite{FFP}, despite spanning 13 scenes, is often treated as a single scene, leading to performance drops when compared to simpler datasets like UCSD Ped2 \cite{T2} and CUHK Avenue \cite{T3}. In real-world applications, such scene and device variations are inevitable, making it impractical to develop specialized models for every setup. To address this, some researchers \cite{UBnormal} have proposed using virtual engines to simulate anomaly events and generate richer positive samples. However, datasets like XD-Violence \cite{HL-Net}, which include movie and game scenes, differ from real-world anomalies, limiting the effectiveness of models trained on them. Bridging this domain gap between virtual and real anomalies is essential for deployable NSVAD systems. Additionally, multimodal NSVAD, while a growing field, remains confined to the fusion of RGB images and synchronized audio, neglecting novel modalities like language \cite{wu2024toward} and texts \cite{pu2024learning}, limiting its application in media streaming and broadcasting.

\subsubsection{Label}
Despite the acknowledged rarity and diversity of anomalies, most models are trained unsupervised \cite{yang2023video,yan2023feature,wang2023video}, using only normal samples. However, as discussed in Section~\ref{sec21}, unsupervised models still require anomaly-free samples during training. Challenges include the impact of pixel noise on model performance and the cost of labeling large-scale datasets for supervised methods. Moreover, UVAD methods rely on data filtering to avoid contaminated training sets, preventing online learning directly from raw video streams, as seen in FuVAD \cite{SDOR}. Conversely, WsVAD utilizes video-level labels to reduce labor costs \cite{MIR}, but improving the stability of FuVAD/UVAD in complex environments or creating hybrid models that can mitigate label noise remain future directions.

\subsubsection{Model}
Unsupervised NSVAD models benefit from self-supervised learning for spatial-temporal feature representation but face challenges with over-generalization, which can lead to missed anomalies \cite{MNAD}. The key challenge is balancing representation and generalization to reduce both false positives and negatives \cite{amp}. Approaches like memory networks and causal representation learning \cite{crc} have shown promise, but performance on complex datasets remains inconsistent. In contrast, WsVAD can only detect predefined anomaly events, limiting its adaptability and failing to meet open-world requirements. While WsVAD models generally offer more reliable results than UVAD and FuVAD, their reliance on ranking loss is still debated. To advance NSVAD, there is a need for interpretable models and systems that integrate data encryption to address privacy concerns.

\subsubsection{System}
Current NSVAD research primarily focuses on improving algorithmic performance on existing datasets, overlooking practical deployment challenges like device heterogeneity and resource constraints. Different applications, such as surveillance analysis, content monitoring, and non-real-time video detection, require distinct task setups and data modalities. However, most work has concentrated on surveillance video detection, neglecting applications like short video analysis and live streaming. Additionally, mobile devices, with their limited storage, computing, and communication resources, necessitate lightweight model designs. Moving forward, NSVAD should focus on the synchronous optimization of detection performance and system cost.

\subsubsection{NSVAD-specialized}
For real-world applications, NSVAD needs to handle vast data streams from urban surveillance cameras while addressing long-term anomaly detection. These systems also face public concerns regarding data overhead and privacy security. Unlike algorithmic research, NSVAD for large-scale applications presents three distinct challenges: (i) Efficient data exchange and task offloading \cite{dong2024task} between millions of cameras and distributed servers. Current research focuses on enhancing algorithms but neglects the costs of data acquisition and transmission, which are critical in large-scale deployments. NSVAD must adopt new video compression \cite{gomes2023video} and transmission protocols to handle the expanding network of cameras and increasing video resolution. Distributed machine learning techniques can also facilitate global model training without aggregating all data. (ii) Detecting anomalies across large spatial and temporal scales. Video IoT systems collect information from entire cities, but current NSVAD models \cite{su2025semantic,zhou2024batchnorm}, designed for discrete scenes, struggle with large-scale data. For example, traffic anomalies like congestion may be easily detected at intersections, but large-scale crowd movements might be misinterpreted as anomalous gatherings. (iii) Privacy protection and ethical concerns. NSVAD systems collect identifiable information such as faces and clothing, raising concerns about privacy breaches and biases related to race, gender, and skin color \cite{su2023prime,su2023prime,noghre2024pheva}. While identity-agnostic data (e.g., encrypted video \cite{cheng2020securead}) and privacy-preserving techniques (e.g., federated learning \cite{al2024collaborative}) are promising, ensuring transparency and fairness remains a significant challenge for NSVAD development.

\subsection{Trends and Opportunities}
Based on the sequential steps of NSVAD depicted in Fig.~\ref{nsvad}, we have summarized the trends and open opportunities from hardware, system, algorithm, and application layers by incorporating the development trends of artificial intelligence and communication technologies, as well as the application requirements of NSVAD in smart cities and mobile internet, aiming to guide researchers in various fields to engage in relevant work.

\subsubsection{Hardware Layer}
The deployment of billions of cameras in roads, factories, and public places not only provides diverse application scenarios but also offers ample data support for training large-scale NSVAD models. Next-generation communication devices and image processing units can provide communication and computational support for building large-scale NSVAD systems, making it feasible for cloud-based global model learning in end-cloud collaborative architecture \cite{ju2024open,teng2024end}. Additionally, emerging sensing devices such as thermal imaging, motion cameras, and event cameras will expand the application potential of NSVAD in scenarios like military and sports. Therefore, we believe that the advancement and innovation of hardware devices strongly drive the development and application of NSVAD technology, and developing deployable NSVAD systems for new types of sensors and large-scale IoT systems will become a trend.

\subsubsection{System Layer}
The system layer aims to bridge the hardware and algorithm layers, providing interfaces for model deployment on terminal devices and supporting the collaborative optimization of computation and communication execution in NSVAD systems. Previous NSVAD research focused on algorithm design while neglecting system layer development for a long time. With the further development of edge artificial intelligence and mobile communication networks, new solutions will be sought for communication strategy optimization, computation offloading, distributed model aggregation, and interface flexibility faced by the system layer, promoting NSVAD towards integration and intelligence.

\subsubsection{Algorithm Layer}
Combining the latest advancements and domain concerns, we believe the development opportunities for algorithms include: 1) the introduction of large-scale heterogeneous datasets; 2) the adaptability transfer of efficient representation learning and reinforcement learning methods \cite{liu2024continual,liu2023efficient,liu2022fast}; 3) innovative combinations of emerging artificial intelligence tasks with NSVAD; and 4) assistance from implicit knowledge in large models \cite{han2024parameter}. Specifically, expanding business scenarios from smart transportation and modern factories provide rich data sources for VSVAD, making it possible to train large-scale models. The rise of online video networks and streaming platforms provides additional data modalities beyond images, such as audio, subtitles, and language, aiding in exploring complex anomaly clues. Furthermore, the development of virtual data engines allows easy emulation of rare anomaly events and provides pixel-level annotations. The rarity and diversity of anomalies and the difficulty of collecting cross-scene videos will no longer be bottlenecks for NSVAD model development. Rich data modalities and volume will drive the development of multimodal and cross-scene NSVAD models and are expected to foster supervised interpretable research routes. The development of deep learning will free NSVAD algorithms from manual features and enable them to extract spatial-temporal features end-to-end and model video normality \cite{zhou2024perceptual,liu2024harmonizing,zhou2024blind,liu2024lcreg}, while the continuous advancement of deep neural networks (such as attention, Transformer, masked autoencoders) \cite{zhou2022quality,liu2022crcnet,zhou2019dual,liu2022few} and emerging representation learning methods (such as video self-supervised learning and causal representation learning) will further improve NSVAD's learning capability. The advancement of emerging artificial intelligence technologies will provide references for addressing specific concerns of NSVAD. For example, unsupervised NSVAD recovers from label-agnostic data biases in complex environments, often encountering significant performance degradation when dealing with diverse common events, a problem long studied in domain generalization tasks. We believe the combination of domain generalization and NSVAD research will help address the negative impact of data bias in unsupervised solutions. Data security and privacy protection have always been concerns for users and researchers when building deployable NSVAD systems for large-scale IoT. Federated learning will provide feasible solutions. The rise of large models will propel generative artificial intelligence research to a climax and give birth to numerous phenomenon-level applications, reshaping industries, and NSVAD is no exception. We believe that large models contain implicit knowledge related to anomalous events, which is crucial for understanding the fundamental differences between normal and abnormal and developing interpretable NSVAD models.

\subsubsection{Application Layer}
On the one hand, applications such as smart transportation and live content monitoring provide ample validation scenarios for NSVAD models, driving the development of cross-scene perspective robust models for real-world deployment. On the other hand, business requirements in specific scenarios will spur new research tasks. For example, thermal imaging, as a completely passive sensing method, has been widely used in autonomous driving and the modern military to overcome the limitations of optical cameras at night. NSVAD model design based on thermal sensing devices will address texture loss and boundary entanglement caused by blackbody radiation, facilitating all-weather anomaly event detection. Additionally, the limited resources of terminal devices and the perceptual range of local systems indicate that lightweight models and the development of end-cloud collaborative NSVAD systems for large-scale applications are worth exploring.

\subsection{Summary}

NSVAD has emerged as a significant area of research with broad implications for smart cities and the mobile internet, exerting essential influence across various domains such as traffic management, industrial manufacturing, and the operations of online video platforms. This influence is critical for maintaining urban safety and ensuring a clear cyberspace. Originating from the confluence of anomaly detection and video understanding, NSVAD has expanded beyond mere algorithm design, transforming into a multifaceted subject of interest that spans AI, IoT, and computing. As the pioneering tutorial-type paper on NSVAD, this article comprehensively outlines its research landscape, clarifying the foundational concepts and developmental trajectories across various research avenues. In particular, we examine recent advancements in unsupervised, weakly supervised, and fully unsupervised methods, providing detailed explanations of classical solutions. Remarkably, this article centrally presents our latest explorations to NSVAD in modern industry, smart cities, and complex systems. Finally, leveraging our experiences, we analyze the challenges, trends, and opportunities within the future vision of NSVAD, aiming to spark inspiration for following engineers and researchers. 
\begin{acks}
  This work was supported in part by the National Natural Science Foundation of China (NSFC) under Grant 62250410368; the Kunshan Municipal Special Project under Grant 24KKSGR024; the Guangdong Pearl River Talent Recruitment Program under Grant 2019ZT08X603; the Guangdong Pearl River Talent Plan under Grant 2019JC01X235; and the National Key Research and Development Program of China under Project No. 2024YFE0200700 (Subject No. 2024YFE0200703). Additional support was provided in part by the Specific Research Fund of the Innovation Platform for Academicians of Hainan Province under Grant YSPTZX202314; the Shanghai Key Research Laboratory of NSAI and the Joint Laboratory on Networked AI Edge Computing, Fudan University-Changan; the China Mobile Research Fund of MOE under Grant KEH2310029; and the NSFC under Grant 62250410368. The authors sincerely thank Liangyu Teng, Yuntian Shi, and Hao Yang from Fudan University for their help in revising this article. Finally, the authors would like to express their gratitude to the anonymous reviewers for their insightful comments and valuable suggestions.
\end{acks}

\bibliographystyle{ACM-Reference-Format}
\bibliography{lib.bib}

\end{document}